\pdfoutput=1%
\documentclass[10pt,twocolumn,letterpaper]{article}

\usepackage[pagenumbers]{cvpr}

\usepackage[T1]{fontenc}
\usepackage{graphicx}
\usepackage{amsmath}
\usepackage{amssymb}
\usepackage{booktabs}
\usepackage{multirow}
\usepackage{colortbl}
\usepackage{xcolor}
\usepackage{enumitem}
\usepackage{xspace}
\usepackage{makecell}

\definecolor{cvprblue}{rgb}{0.21,0.49,0.74}
\colorlet{eccvblue}{cvprblue}
\usepackage{xr-hyper}
\usepackage[pagebackref,breaklinks,colorlinks,allcolors=cvprblue]{hyperref}
\usepackage[capitalize]{cleveref}

\def\confName{CVPR}
\def\confYear{2027}

\newcommand{\ourmodel}{DeltaRAE\xspace}

\newcommand{\safeincludegraphics}[2][width=\linewidth]{%
  \IfFileExists{#2}{%
    \includegraphics[#1]{#2}%
  }{%
    \fbox{\parbox[c][4.2cm][c]{0.96\linewidth}{\centering\small\textbf{[Figure placeholder: \texttt{\detokenize{#2}}]}}}%
  }%
}

\usepackage{placeins}
\usepackage{tabularx}
\newcolumntype{L}{>{\raggedright\arraybackslash}X}
\newcolumntype{C}{>{\centering\arraybackslash}X}
\usepackage{docmute}

\title{DeltaRAE: Derivative Tokenization for Long Video Generation}

\author{%
Mingzhuo Yu$^{1}$ \quad Shengcao Cao$^{2}$ \quad Chengzhi Mao$^{3}$ \quad Yong Cheng$^{2}$ \quad Di Jiang$^{1}$ \quad Lijun Yu$^{2}$\\[3pt]
$^{1}$Google \qquad $^{2}$Google DeepMind \qquad $^{3}$Rutgers University%
}

\begin{document}

\twocolumn[{%
\renewcommand\twocolumn[1][]{#1}%
\maketitle
\begin{center}
    \captionsetup{type=figure}
    \includegraphics[width=\textwidth]{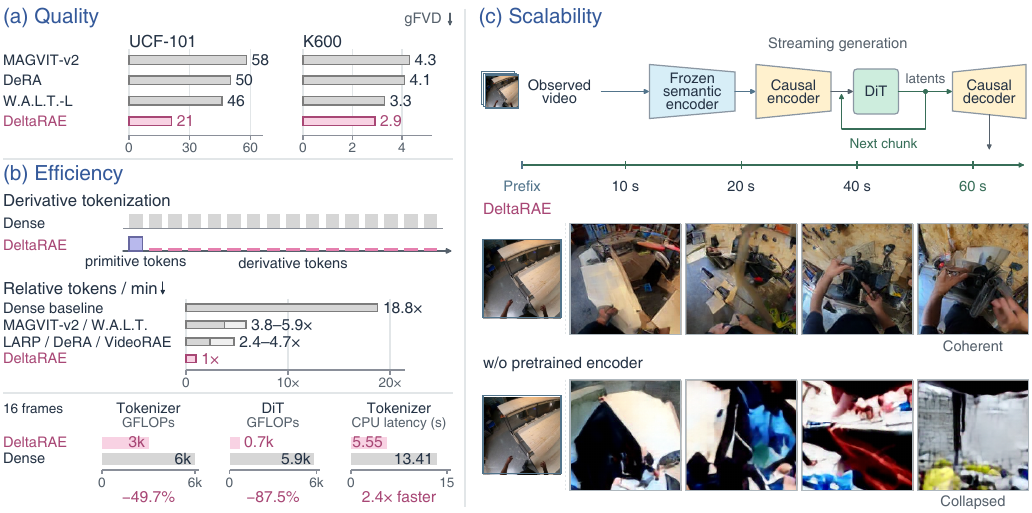}
    \captionof{figure}{\textbf{Overview of \ourmodel, a causal representation autoencoder for video.} \textbf{(a)} A diffusion transformer trained on its latents reaches the lowest gFVD among the listed tokenizers on UCF-101 and Kinetics-600. \textbf{(b)} Derivative tokenization gives periodic anchor frames a full set of \emph{primitive tokens} and all remaining frames a few \emph{derivative tokens}. A minute of video takes $18.8\times$ fewer tokens than dense per-frame features, and a 16-frame clip about half the tokenizer GFLOPs. \textbf{(c)} \ourmodel scales to one-minute generation: its 60-second rollout stays coherent, and the rollout without the pretrained encoder collapses.}
    \label{fig:teaser_overview}
\end{center}%
\vspace{4pt}
}]

\begin{abstract}
Latent video generators are trained on clips of a few seconds, and much of this limit lies in their tokenizers, which spend as many tokens on a static frame as on a moving one and need future frames that do not exist during generation. We introduce \ourmodel, a causal representation autoencoder for video. Its derivative tokenization encodes periodic anchor frames in full and all remaining frames with a few tokens each, and its causal encoder and decoder read a frozen pretrained image encoder. On UCF-101 and Kinetics-600, a standard diffusion transformer trained on \ourmodel latents is competitive with state-of-the-art video tokenizers. \ourmodel halves the tokenizer GFLOPs on 16-frame clips relative to keeping the same number of patch tokens for every frame, and it represents a minute of video with $18.8\times$ fewer tokens. The tokenizer trains end to end on one-minute videos, and in video prediction a generator on its latents produces a full minute of video. The same generator collapses within 15 seconds in most videos when the tokenizer has the same token budget but no pretrained encoder.
\end{abstract}

\section{Introduction}
\label{sec:intro}
Latent video generators have grown from convolutional U-Nets to large diffusion transformers~\cite{peebles2023scalable,esser2024scaling,gupta2024walt}, yet they are still trained on clips of a few seconds. Generating a minute of video with one model needs a latent sequence that is short enough to train on and that can be produced in temporal order. The limit is set less by the generator than by the tokenizer in front of it.

Existing tokenizers stand in the way. Most of them map video onto a uniform spatiotemporal grid~\cite{yan2021videogpt,yu2023magvit,yu2024magvitv2,wang2024omnitokenizer} and spend the same number of tokens on every frame, whether or not anything moves, so long videos are produced by stitching short clips. Global 1D tokenizers~\cite{wang2025larp,tan2025sweettok,guo2025dera,zhong2025vfrtok} use fewer tokens per clip, but every token depends on the whole clip, so a prefix cannot be encoded on its own and no frame can be decoded before the clip is complete. Latents trained only to reconstruct pixels can also be hard to generate. Representation autoencoders~\cite{zheng2025diffusion,xie2026videorae} avoid this by decoding from a frozen pretrained encoder, but they keep a uniform token budget and clip-level context.

We present \ourmodel, a causal representation autoencoder for video (\cref{fig:teaser_overview}). Consecutive frames share most of their content, so we treat tokens the way a video codec treats frames: periodic anchor frames are encoded in full, and every frame in between receives a few tokens that describe its change from the past. We call this \emph{derivative tokenization}. The tokens are read from a frozen semantic encoder, DINOv2 in our experiments, by a causal encoder, and a causal decoder maps them back to pixels, so neither the tokens nor the reconstruction of a prefix depend on later frames. A video can therefore be generated chunk by chunk, entirely in latent space, which we call streaming generation.

A generator trained on 10-second chunks of \ourmodel latents produces a full minute of video that stays coherent in all of 455 held-out videos. The same generator collapses in nearly all of them on \ourmodel without the pretrained encoder, which has the same token budget and reconstructs better than ours (\cref{fig:long_gen}). A diffusion transformer trained on \ourmodel latents is also competitive at 16 frames, with a gFVD of 21 on UCF-101 at 736 tokens and 2.9 on Kinetics-600 frame prediction at 1,216 (\cref{fig:tokens_gfvd,tab:main}), and our ablations show the same gap between reconstruction and generation there (\cref{sec:analysis}).

Our contributions are the following.
\begin{itemize}[leftmargin=*,nosep]
    \item \emph{Derivative tokenization}, which gives periodic anchor frames a full set of tokens and all remaining frames a few. A minute of video becomes 19,584 tokens, $18.8\times$ fewer than the 368,640 obtained by keeping every patch token of every frame, which makes one-minute generation feasible in cost.
    \item \emph{A causal representation autoencoder for video}, trained end to end on one-minute videos. Its tokens are read causally from a frozen semantic encoder, so a video is generated chunk by chunk in latent space, with one tokenizer and one generator for generation and prediction.
    \item \emph{State-of-the-art generation quality.} On UCF-101 and Kinetics-600, \ourmodel reaches the lowest gFVD among the tokenizers of \cref{tab:main} with fewer tokens and half the tokenizer GFLOPs.
\end{itemize}

\section{Related Work}
\label{sec:related}

\noindent\textbf{Video tokenizers.}
Discrete~\cite{van2017neural,yu2022vector,yan2021videogpt,ge2022long,yu2023magvit,yu2024magvitv2,wang2024omnitokenizer} and continuous~\cite{rombach2022high,gupta2024walt,hansen2025vitok} tokenizers compress video with 3D convolutions or spatiotemporal patches at a fixed stride and assign the same number of tokens to every latent frame. MAGVIT-v2~\cite{yu2024magvitv2} and W.A.L.T.~\cite{gupta2024walt} use a temporally causal 3D CNN so that the first frame can be encoded as an image, but keep a uniform grid. Recent 1D tokenizers drop the grid, as TiTok~\cite{yu2024image} did for images: LARP~\cite{wang2025larp}, SweetTok~\cite{tan2025sweettok}, DeRA~\cite{guo2025dera} and VFRTok~\cite{zhong2025vfrtok} compress a fixed-length clip with queries that attend to all frames. Each token then depends on the entire clip, so an isolated prefix cannot be encoded consistently and longer videos need stitching. Closest to ours in structure is AdapTok~\cite{li2025adaptok}, a causal 1D tokenizer that gives each four-frame block a content-dependent number of discrete tokens learned from pixels, trained and evaluated on 16-frame clips; EVATok~\cite{xiong2026evatok} predicts this allocation with a router, and ElasticTok~\cite{yan2025elastictok}, InfoTok~\cite{ye2026infotok} and VideoFlexTok~\cite{atanov2026videoflextok} also adapt the token count to content. \ourmodel reads its tokens from a frozen semantic encoder, keeps them continuous, and is trained on one-minute videos.

\noindent\textbf{Pretrained representations for generation.}
Aligning diffusion features with self-supervised encoders speeds up training~\cite{yu2025representation}, and RAE~\cite{zheng2025diffusion} shows that an image generator can operate directly in the feature space of a frozen DINOv2~\cite{oquab2023dinov2} or SigLIP~2~\cite{zhai2023sigmoid,tschannen2025siglip} encoder. VideoRAE~\cite{xie2026videorae} extends RAE to video by compressing a frozen video foundation model into continuous or discrete latents for a fixed-length clip. The concurrent TrackTok~\cite{pham2026tracktok} builds object-aligned 1D tokens on a frozen video encoder. \ourmodel shares the premise that a frozen encoder gives a latent space that is easier to generate. It differs in reading a per-frame image encoder through causal registers, so that latents never depend on future frames, in letting the token budget vary over time, and in targeting one-minute streaming generation rather than short clips.

\noindent\textbf{Predictive coding and long video.}
Video codecs~\cite{sullivan2012overview} and neural codecs~\cite{lu2019dvc,hu2021fvc} separate intra-coded frames from predicted frames inside groups of pictures. \ourmodel applies the same split to the latent tokens of a generative model. iVideoGPT~\cite{wu2024ivideogpt} and Reducio~\cite{tian2025reducio} likewise encode a context frame in full and later frames with few tokens conditioned on it, and DeltaTok~\cite{kerssies2026deltatok} compresses the change between consecutive frames of frozen DINOv3 features into one token for dense forecasting, without a pixel decoder. \ourmodel adds periodic anchors and a causal pixel decoder, and targets generation. Long-video generators roll a short-clip model over sliding windows~\cite{ge2022long,hong2023cogvideo,chen2024diffusion}, compress the frame history~\cite{zhang2025framepack}, add test-time-training layers~\cite{dalal2025one} or plan in a semantic space~\cite{bai2025semanticgen}; CoordTok~\cite{jang2025coordtok} tokenizes 128-frame clips as triplanes. Our contribution is a causal latent sequence cheap enough to train on a full minute, paired with a standard rectified-flow~\cite{liu2023flow,lipman2023flow,esser2024scaling} transformer.

\section{Method}
\label{sec:method}

\begin{figure*}[t]
    \centering
    \includegraphics[width=\textwidth]{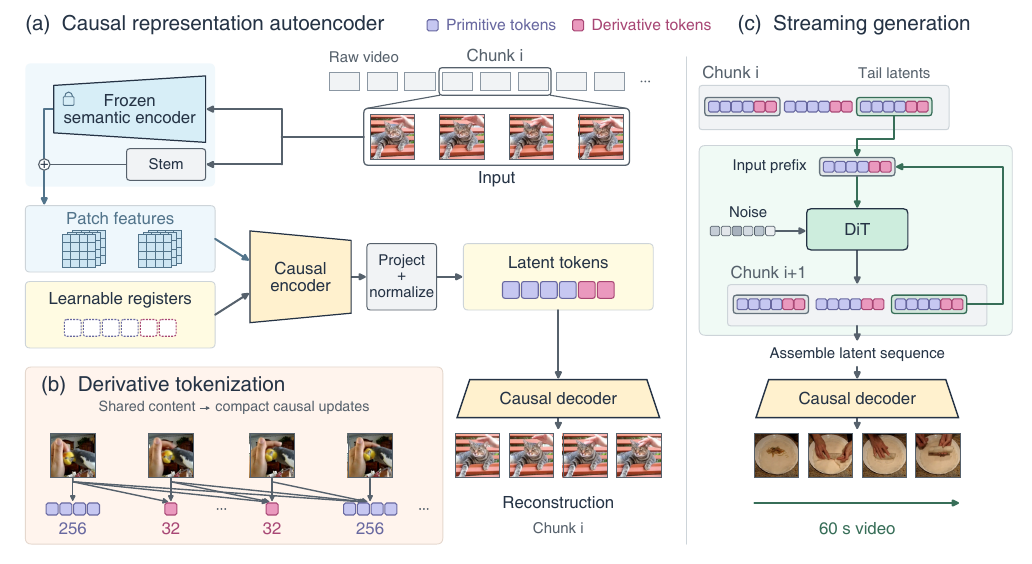}
    \caption{\textbf{Architecture of \ourmodel.} \textbf{(a) Causal representation autoencoder.} A causal encoder updates learnable registers from frozen semantic features augmented by a pixel-residual stem. The updated registers are projected and normalized into latent tokens for causal decoding. \textbf{(b) Derivative tokenization.} To exploit shared content over time, periodic anchor groups receive a full set of primitive tokens, while the remaining groups use fewer derivative tokens with earlier groups as context. \textbf{(c) Streaming generation.} A DiT generates successive chunks, reusing each chunk's tail latents as the next input prefix. The assembled latent sequence is decoded into a 60-second video.}
    \label{fig:detailed_arch}
\end{figure*}

\subsection{Derivative tokenization}
\label{sec:formulation}

A tokenizer for long video has to decide how many tokens each moment receives. We give a full set to a few frames and a small set to all others, and define this schedule over groups of frames so that one rule covers short clips and one-minute videos. A group is the unit that receives tokens: a single frame in a short clip, and four consecutive frames in a one-minute video. Given a video $X\in\mathbb{R}^{T\times H\times W\times 3}$, we split time into $G=T/\Delta t$ groups of $\Delta t$ frames. Short clips of $T=16$ frames use $\Delta t=1$, and one-minute videos of $T=1{,}440$ frames use $\Delta t=4$, which gives $G=360$ groups. Group $g$ is an \emph{anchor} if $g \bmod a = 0$ for an anchor period $a$, and a \emph{derivative} group otherwise (\cref{fig:detailed_arch}b). The number of latent tokens assigned to group $g$ is
\begin{equation}
N(g)=\begin{cases}
N_a = 256, & g \bmod a = 0 \quad\text{(primitive tokens)},\\
N_d, & \text{otherwise} \quad\text{(derivative tokens)},
\end{cases}
\label{eq:token_schedule}
\end{equation}
with $N_d\in\{32,64\}$. A 16-frame clip with a single anchor has $256+15N_d$ tokens, that is 736 or 1,216, compared with 4,096 for a dense per-frame grid. We refer to each variant by this token count. A token is a continuous vector, so the count measures the length of the sequence that the generator sees and not a bit rate. Nothing in the loss forces derivative tokens to encode only what changed, but their small number and their access to earlier groups favor it. A one-minute video with an anchor every 40 frames, $a=10$, and $N_d=32$ has $36\,(256+9\cdot 32)=19{,}584$ tokens, or 13.6 per frame.

\subsection{Causal representation encoder}
\label{sec:semantic_compression}

\noindent\textbf{Frozen semantic features.}
The encoder turns the dense features of a frozen image model into a few tokens per group, using learnable registers~\cite{darcet2024vision} that read the features under a causal mask and then serve as the latent tokens (\cref{fig:detailed_arch}a). The image model is a frozen semantic encoder. All experiments use DINOv2 ViT-B/14~\cite{oquab2023dinov2} at $224^2$, which gives $256$ patch features of width 768 per frame; SigLIP~2 can serve as the backbone as well (\cref{sec:supp_recon}). For $\Delta t=4$, the features of the four frames in a group are averaged, and a zero-initialized linear layer over their concatenation is added so that the model can learn to keep within-group motion. We write $F_g\in\mathbb{R}^{256\times 768}$ for the features of group $g$.

\noindent\textbf{Registers under a causal mask.}
For every group we take $N(g)$ learnable register tokens $Q_g$ from a shared bank, concatenate $[F_g, Q_g]$ over all groups, and apply a lightweight transformer whose attention mask lets a token of group $g$ attend to a token of group $g'$ only if
\begin{equation}
\lfloor g/C\rfloor=\lfloor g'/C\rfloor \;\;\text{and}\;\; g'\le g ,
\label{eq:block_causal_mask}
\end{equation}
where $C$ is the chunk length, 16 groups for clips and 60 groups, or 240 frames, for long video. Spatial position is encoded by a fixed 2D sine-cosine embedding added to the patch features, and temporal position by a rotary embedding~\cite{su2021roformer} over the group index. The updated patch features are discarded. Each updated register $h_{g,i}$ is projected by a linear map $W$ and normalized,
\begin{equation}
z_{g,i}=\frac{W h_{g,i}}{\lVert W h_{g,i}\rVert_2}.
\label{eq:unit_sphere}
\end{equation}
The unit-sphere latent has a fixed scale without a codebook or a KL term, which keeps the generator's noise schedule independent of the token budget.

\noindent\textbf{Pixel-residual stem.}
DINOv2 features discard sub-patch position and color detail. A decoder that sees the whole clip can partly recover it from neighboring frames, but a causal decoder cannot. We therefore add to $F_g$ the output of a small convolutional stem applied to the concatenation of the frame and its difference to the previous frame, $[X_g,\; X_g-X_{g-1}]$, where the difference is zero for the first frame. The last layer of the stem is zero-initialized, so training starts from the model without the stem. It uses only the current and the previous frame, so the encoder stays causal.

\subsection{Causal decoder and prefix invariance}
\label{sec:asymmetric_causality}

The decoder is a deeper transformer over the latent tokens and a set of learnable target tokens, $256$ per group for anchor and derivative groups alike. Its attention mask is chosen so that the reconstruction of group $g$ depends only on latent tokens of groups $g'\le g$ in the same chunk. Target tokens are mapped linearly to pixel patches. The full target grid costs no latent tokens and matters for quality (\cref{tab:long_design_ablation_supp}).

With \cref{eq:block_causal_mask} in the encoder and the causal decoder, both the latents and the reconstruction of a prefix of groups are independent of what follows in the chunk:
\begin{equation}
\mathrm{Enc}(X_{1:g_0})_{g}=\mathrm{Enc}(X_{1:G})_{g},\quad
\mathrm{Dec}(Z_{1:g_0})_{g}=\mathrm{Dec}(Z_{1:G})_{g}
\label{eq:prefix_invariance}
\end{equation}
for all $g\le g_0\le G$. We verify this numerically: replacing the frames after a prefix of 4, 8 or 12 frames with noise changes prefix latents and prefix reconstructions by exactly zero (\cref{sec:supp_causality}). One tokenizer and one generator therefore serve unconditional generation and prediction from a prefix of any number of groups. Continuation beyond a chunk reuses its last latents as the prefix of the next (\cref{sec:diffusion_fcfg}).

\subsection{Scaling to one-minute videos}
\label{sec:long}

The design extends from 16 to 1,440 frames through periodic anchors, chunked causality and two-stage training. \emph{Periodic anchors.} With an anchor every $a=10$ groups, that is every 40 frames or 1.67 s, no derivative group is more than nine groups away from a full description of the scene. \emph{Chunked causality.} Because the masks are block-diagonal over chunks of 240 frames, attention spans at most 10 seconds, and encoding or decoding a chunk in isolation gives the same result as processing it inside the full video. The tokenizer is trained on full one-minute videos with sequence-parallel attention that skips the masked blocks, so the step time grows linearly with duration (\cref{sec:supp_impl}), and the generator is trained on single chunks. \emph{Two-stage training.} The tokenizer is first trained with a pixel $L_2$ loss. The encoder is then frozen, and the causal decoder is finetuned with perceptual, adversarial and temporal-difference losses. The latent space seen by the generator is fixed after the first stage.

\noindent\textbf{Training objective.}
All tokenizers minimize
\begin{equation}
\mathcal{L}=\lambda_{\text{rec}}\lVert X-\hat X\rVert_2^2+\lambda_{\text{perc}}\mathcal{L}_{\text{LPIPS}}+\lambda_{\text{adv}}\mathcal{L}_{\text{GAN}},
\label{eq:tok_loss}
\end{equation}
with the LPIPS perceptual loss~\cite{zhang2018unreasonable}, a patch discriminator and hinge loss~\cite{yu2022vector,yu2023magvit}. Short-clip tokenizers start from a single-frame tokenizer trained on ImageNet~\cite{deng2009imagenet} and are then extended in time (\cref{sec:supp_protocol}).

\subsection{Streaming generation}
\label{sec:diffusion_fcfg}

We train a diffusion transformer, DiT~\cite{peebles2023scalable}, on the flattened latent sequence with a rectified-flow objective~\cite{liu2023flow,esser2024scaling} and sample with 100 Euler steps by default. For prediction and continuation, the latents of the observed prefix are given clean: they replace the noisy tokens at prefix positions, a binary mask is concatenated to the input, and the timestep embedding at prefix positions is set to the clean end of the flow. Positional embeddings are factorized into frame index, slot index within the frame, and token role, primitive or derivative. We use query-key normalization in attention and an exponential moving average of the weights. Generator configurations are listed in \cref{sec:supp_protocol}.

\noindent\textbf{Chunk-wise streaming.}
For one-minute generation the model is trained on 240-frame chunks of 3,264 tokens, conditioned on the first 40 frames, which hold 544 tokens. At inference each new chunk is conditioned on the last 40 frames of the previous one (\cref{fig:detailed_arch}c), so seven chunks cover $40+7\cdot 200=1{,}440$ frames. These ten groups begin with an anchor and keep their token roles when they move to the start of the new chunk. Because the tokenizer is causal, their generated latents are final. We copy them to the start of the new chunk and re-project them to the unit sphere, so no frame is decoded to pixels and encoded again at a chunk boundary. The copied latents were formed with the earlier groups of their chunk as context, a shift that \cref{eq:prefix_invariance} does not cover. We therefore add Gaussian noise to the prefix latents during training and validate this step empirically (\cref{sec:supp_method}).

\section{Experiments}
\label{sec:experiments}

\subsection{Setup}
\label{sec:setup}

\noindent\textbf{Datasets and tasks.}
We use Kinetics-600~\cite{carreira2018short} for reconstruction and for frame prediction with 5 conditioning frames and 11 predicted frames, UCF-101~\cite{soomro2012ucf101} for reconstruction and class-conditional generation, and one-minute clips from Ego4D~\cite{grauman2022ego4d} for long-video reconstruction and continuation, with 9,935 training videos and held-out videos that share no participant with training. One-minute reconstruction is also evaluated zero-shot on 256 YouCook2~\cite{zhou2018towards} validation videos, which are not seen in training. Results on Something-Something-v2~\cite{goyal2017something} are in the appendix.

\noindent\textbf{Metrics.}
We report reconstruction and generation FVD~\cite{unterthiner2019fvd} for 16-frame benchmarks, together with PSNR, SSIM~\cite{wang2004image} and LPIPS. rFVD compares decoded videos with their inputs on 50,000 clips, and gFVD compares newly generated videos with real ones. Kinetics-600 frame prediction follows the protocol of MAGVIT~\cite{yu2023magvit}: 50,000 clips from the validation split, 2 clips $\times$ 3 crops per video, evaluated at $64^2$. UCF-101 gFVD is computed between 10,000 class-conditional samples and the training split at $128^2$, as in MAGVIT, LARP and AdapTok. Ablation tables state their own sample count. For one-minute videos we report reconstruction PSNR, SSIM, and the drift, which is the slope of a linear fit of the mean PSNR over the minute. One-minute generation is evaluated by rollouts, minutes of video generated chunk by chunk from a prefix, with the same generator recipe for every tokenizer (\cref{tab:long_cap,fig:long_gen}). Numbers of prior work are taken from the respective papers.

\noindent\textbf{Implementation.}
The register encoder is lightweight and the decoder is deeper; layer counts are listed in \cref{sec:supp_impl}. Tokenizers with the pixel-residual stem are finetuned from earlier tokenizers of the same token budget. Tokenizers and generators are trained on Kinetics-600 and UCF-101. Short-clip and one-minute models share the architecture and are trained separately. Kinetics-600 reconstruction numbers use EMA weights. Schedules, initialization and checkpoint choices are listed in \cref{sec:supp_protocol}.

\subsection{Quality: short-video benchmarks}
\label{sec:exp_main}

\begin{table*}[t]
    \centering
    \small
    \caption{\textbf{Comparison with prior tokenizers on 16-frame benchmarks.} UCF-101: rFVD on 50,000 clips and class-conditional gFVD with 10,000 samples against the training split. \ourmodel operates at $224^2$, and its outputs are evaluated at $128^2$ and at $256^2$ to match the methods of each block. K600: frame-prediction gFVD at $64^2$ on 50,000 clips. $^\ast$Uses extra training data.}
    \label{tab:main}
    \setlength{\tabcolsep}{5pt}
    \begin{tabular}{l c c | c c | c}
        \toprule
        & & & \multicolumn{2}{c|}{\textbf{UCF-101}} & \textbf{K600 pred.} \\
        \textbf{Method} & \textbf{Res.} & \textbf{Tokens / 16f} & \textbf{rFVD} $\downarrow$ & \textbf{gFVD} $\downarrow$ & \textbf{gFVD} $\downarrow$ \\
        \midrule
        TATS~\cite{ge2022long} & $128^2$ & 1,024 & 162 & 332 & --- \\
        OmniTokenizer~\cite{wang2024omnitokenizer} & $128^2$ & 1,280 & 42 & 191 & 32.9 \\
        MAGVIT~\cite{yu2023magvit} & $128^2$ & 1,024 & 25 & 76 & 9.9 \\
        AdapTok~\cite{li2025adaptok} & $128^2$ & 2,048 & 28 & 67 & 11 \\
        MAGVIT-v2~\cite{yu2024magvitv2} & $128^2$ & 1,280 & 8.62 & 58 & 4.3 \\
        LARP~\cite{wang2025larp} & $128^2$ & 1,024 & 20 & 57 & 5.1 \\
        DeRA~\cite{guo2025dera} & $128^2$ & 1,024 & 15 & 50 & 4.1 \\
        W.A.L.T.-L~\cite{gupta2024walt} & $128^2$ & 1,280 & --- & 46 & 3.3 \\
        ViTok$^\ast$~\cite{hansen2025vitok} & $128^2$ & 1,024 & 8.04 & 27 & --- \\
        \rowcolor{cvprblue!10}\ourmodel\ (Ours) & $128^2$ & 1,216 & \textbf{5.06} & 26 & \textbf{2.9} \\
        \rowcolor{cvprblue!10}\ourmodel\ (Ours) & $128^2$ & \textbf{736} & 8.12 & \textbf{21} & 5.2 \\
        \midrule
        VFRTok-L~\cite{zhong2025vfrtok} & $256^2$ & 512 & 13.8 & 71 & --- \\
        SweetTok~\cite{tan2025sweettok} & $256^2$ & 1,280 & 20 & 65 & --- \\
        VideoRAE~\cite{xie2026videorae} & $256^2$ & 1,024 & 13 & 40 & --- \\
        \rowcolor{cvprblue!10}\ourmodel\ (Ours) & $256^2$ & 1,216 & \textbf{5.47} & 27 & --- \\
        \rowcolor{cvprblue!10}\ourmodel\ (Ours) & $256^2$ & 736 & 8.69 & \textbf{22} & --- \\
        \bottomrule
    \end{tabular}
\end{table*}

\begin{figure}[t]
    \centering
    \includegraphics[width=\linewidth]{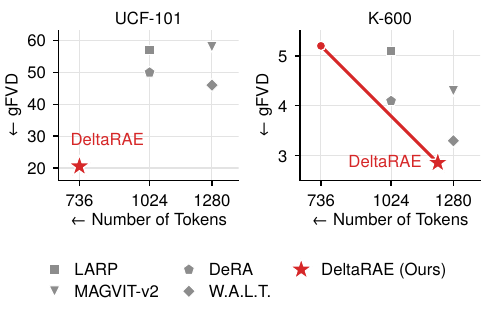}
    \caption{\textbf{Generation quality against token count.} gFVD against tokens per 16 frames, with the numbers of \cref{tab:main}. UCF-101 shows the 736-token model. TATS, OmniTokenizer, MAGVIT and AdapTok lie outside the plotted range, and ViTok (external training data) is not plotted.}
    \label{fig:tokens_gfvd}
\end{figure}

\begin{figure}[t]
    \centering
    {\sffamily\let\small\footnotesize\input{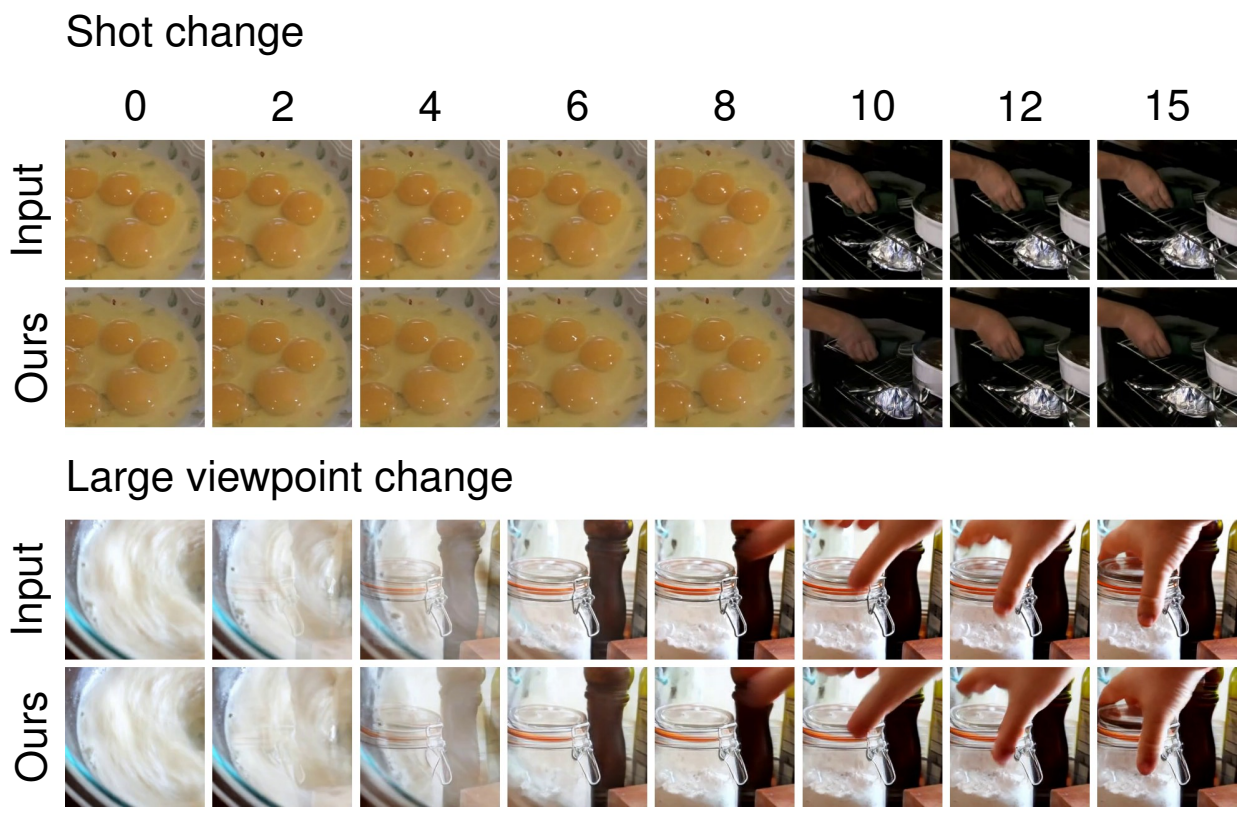}}%
    \caption{\textbf{Reconstruction of difficult 16-frame clips with 736 tokens.} Input and reconstruction by \ourmodel on two held-out clips, with frame indices on top. Every frame after the first has 32 derivative tokens, including the frames after the shot change. \cref{fig:supp_edge_cases} shows more clips.}
    \label{fig:edge_cases}
\end{figure}

We compare with prior tokenizers in \cref{tab:main} and \cref{fig:tokens_gfvd}. On UCF-101, \ourmodel reaches a gFVD of 21 with 736 tokens, and 22 at $256^2$. For Kinetics-600 frame prediction, 1,216 tokens give 2.9. All three are the lowest values in their block of \cref{tab:main}. The 736-token UCF-101 generator is trained for 12,000 epochs and the 1,216-token generator for 1,500, which gives 26 (\cref{tab:arch_dit_supp}); with the same generator the two budgets are on par (\cref{sec:supp_gen}). VideoRAE~\cite{xie2026videorae} also decodes from a frozen pretrained encoder, and \cref{tab:main} lists its discrete tokenizer. Its continuous latents are evaluated with 2,048 samples; under that protocol \ourmodel scores 82, below the values it reports (\cref{tab:videorae}; protocols in \cref{sec:supp_protocol}). \cref{fig:edge_cases} shows reconstructions of two difficult clips.

\subsection{Efficiency}
\label{sec:exp_efficiency}

\begin{table}[t]
    \centering
    \small
    \caption{\textbf{Compute for 16 frames}, \ourmodel with 736 tokens compared with 4,096 dense per-frame tokens, measured from compiled graphs. Tokenizer cost covers encoder and decoder, with latency on CPU. DiT GFLOPs are for one forward pass, and DiT latency is for one sample with 250 steps.}
    \label{tab:compute_cost}
    \setlength{\tabcolsep}{4pt}
    \begin{tabular}{@{}l c c c@{}}
        \toprule
        & \textbf{\ourmodel\ (Ours)} & \textbf{Dense} & \textbf{Change} \\
        \midrule
        Tokenizer GFLOPs & \textbf{3,015} & 5,999 & $-49.7\%$ \\
        Tokenizer latency (s) & \textbf{5.55} & 13.41 & $2.4\times$ faster \\
        DiT GFLOPs & \textbf{730} & 5,859 & $-87.5\%$ \\
        DiT latency (s) & \textbf{10.0} & 16.6 & $1.66\times$ faster \\
        \bottomrule
    \end{tabular}
\end{table}

Reducing 4,096 dense tokens to 736 halves the tokenizer cost and lowers the cost of a DiT forward pass by $87.5\%$ and sampling latency by $40\%$ (\cref{tab:compute_cost}), with no loss in generation quality when tokenizer and generator are trained with the same recipe (\cref{tab:ucf_gen_supp}). One minute of video takes 19,584 tokens instead of 368,640 for dense per-frame features, a factor of $18.8$, of which 4 comes from grouping four frames and 4.7 from the anchor and derivative budget. A 10-second chunk then holds 3,264 tokens, short enough to train a DiT on.

\subsection{Scalability: one-minute videos}
\label{sec:exp_long}

\begin{figure*}[t]
    \centering
    \includegraphics[width=0.75\textwidth]{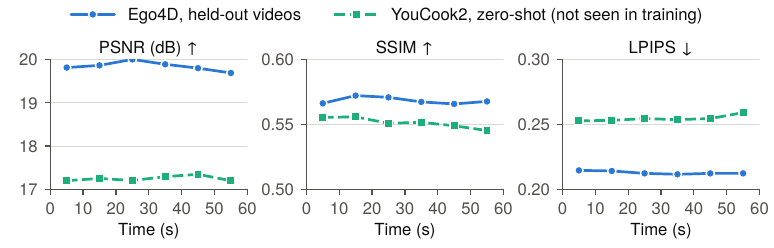}
    \caption{\textbf{Reconstruction of one-minute videos over time.} PSNR, SSIM and LPIPS of the one-minute \ourmodel tokenizer over 1,440 frames, averaged over each 10-second segment, on held-out Ego4D videos and, zero-shot, on YouCook2 videos not seen in training. Quality is stable over the minute on both.}
    \label{fig:recon_curves}
\end{figure*}

\noindent\textbf{Reconstruction.}
The reconstruction quality of the one-minute tokenizer stays flat over time (\cref{fig:recon_curves}), with a fitted PSNR slope of $-0.10$ dB per minute and no drop at the boundaries of the 10-second chunks. In the zero-shot evaluation on YouCook2, third-person cooking videos not seen in training, it reaches 17.25 dB PSNR, 0.551 SSIM and 0.255 LPIPS, compared with 19.84 dB, 0.568 and 0.213 on Ego4D, and stays as flat, at $+0.07$ dB per minute. We compare with \ourmodel without the pretrained encoder, which keeps the token schedule and the causal masks but reads pixel patches in place of the frozen encoder. At 13.6 tokens per frame \ourmodel is less faithful than this variant (19.84 vs.\ 23.29 dB PSNR) and than 16-frame tokenizers applied clip by clip with $3.4\times$ to $5.6\times$ as many tokens (\cref{tab:long_recon}). The latents of the variant reconstruct better but, as shown next, do not survive generation.

\begin{table*}[t]
    \centering
    \small
    \caption{\textbf{One minute of video, $1{,}440$ frames, under different tokenizers.} The published tokenizers (top) are an image RAE applied to every frame, tokenizers that compress each 16-frame clip into one token sequence, and 3D-convolutional VAEs with a uniform latent grid. They need several times more tokens per minute than \ourmodel and are not trained at this length. With the same $19{,}584$ tokens and the same generator recipe (bottom), 60-second rollouts stay coherent with the frozen DINOv2 encoder and collapse in nearly all videos without it; a rollout counts as coherent if its scene does not break down into distorted textures.}
    \label{tab:long_cap}
    \setlength{\tabcolsep}{7pt}
    \begin{tabular}{@{}l c c c c c@{}}
        \toprule
        \textbf{Tokenizer} & \textbf{Tokens / min} & \textbf{Trained on} & \textbf{Causal} & \textbf{Pretrained encoder} & \textbf{Coherent rollouts} \\
        \midrule
        RAE~\cite{zheng2025diffusion} on every frame & $368{,}640$ & 1 frame & \checkmark & DINOv2 & --- \\
        LARP, DeRA, VideoRAE~\cite{wang2025larp,guo2025dera,xie2026videorae} & $46{,}080$--$92{,}160^\dagger$ & 16 frames & $\times$ & none or video & --- \\
        MAGVIT-v2, W.A.L.T.~\cite{yu2024magvitv2,gupta2024walt} & $73{,}728$--$115{,}200$ & 17 frames & \checkmark & none & --- \\
        \midrule
        \ourmodel w/o pretrained encoder & $19{,}584$ & $1{,}440$ frames & \checkmark & none & 2 of 455 \\
        \rowcolor{cvprblue!10}\textbf{\ourmodel\ (Ours)} & $\mathbf{19{,}584}$ & $1{,}440$ frames & \checkmark & \textbf{DINOv2} & \textbf{455 of 455} \\
        \bottomrule
    \end{tabular}

    \vspace{2pt}
    {\footnotesize $^\dagger$Token count if $90$ independent $16$-frame clips of $512$ to $1{,}024$ tokens each are stitched across one minute.}
\end{table*}

\begin{figure*}[t]
    \centering
    \scalebox{0.90}{\sffamily\input{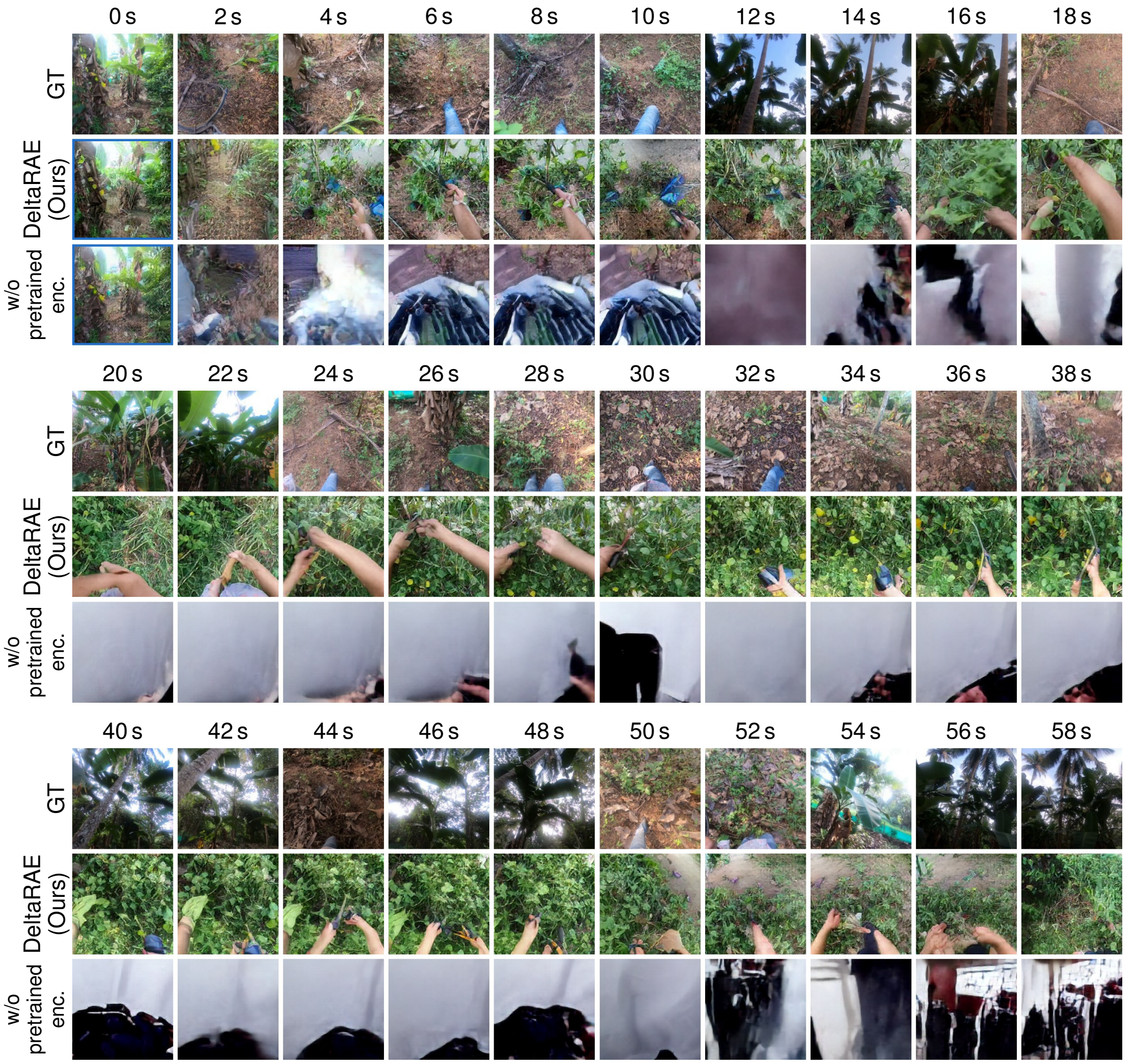}}
    \caption{\textbf{One-minute continuation from a 1.67-second prefix on a held-out Ego4D video}, one frame every 2 seconds. Each block covers 20 seconds with the ground truth (GT), \ourmodel, and \ourmodel without the pretrained encoder with the same 19{,}584 tokens per minute and generator recipe. The framed frame is the prefix. \cref{fig:supp_long_gen_full,fig:supp_long_gen} show more videos and a 20-second prefix.}
    \label{fig:long_gen}
\end{figure*}

\noindent\textbf{Generation.}
We compare tokenizer families at 1,440 frames in \cref{tab:long_cap}. The tokenizer is trained end to end on 1,440 frames. Because a minute is only 19,584 tokens, several times fewer than for other tokenizer families (\cref{fig:teaser_overview}b), a single DiT trained on 10-second chunks is affordable and generates 58.3 seconds from a 1.67-second prefix in seven chunks, without decoding to pixels between chunks. We generate one-minute rollouts with both tokenizers for 455 held-out videos and call a rollout coherent if its scene does not break down into distorted textures. \ourmodel stays coherent in all 455. The variant without the pretrained encoder, with the same token budget and generator recipe, collapses in nearly all of them. For 27 of the 455 we rendered the rollouts in full: the variant collapses within the first 15 seconds in 22. \ourmodel keeps the scene layout through 60 seconds in 25 (\cref{fig:long_gen}), and in the other two the picture fades to black in the second half of the minute. \cref{fig:supp_long_gen} shows continuations from a 20-second prefix. A 16-frame comparison with a tokenizer without the pretrained encoder at matched rFVD, run with a simpler DDPM~\cite{ho2020denoising} generator, points the same way (\cref{sec:supp_fr4}).

\section{Analysis}
\label{sec:analysis}

\subsection{Causality in the encoder and the decoder}
\label{sec:ana_causality}

We train all four combinations of a bidirectional or causal encoder with a bidirectional or causal decoder under one schedule (\cref{tab:causality}). \emph{Reconstruction and generation rank the designs differently.} The fully bidirectional tokenizer has the lowest rFVD on both datasets and the highest gFVD of the four, 31. A causal encoder with a bidirectional decoder has the lowest, 26. \emph{A causal decoder costs quality, and the stem recovers it.} Making the decoder causal as well raises gFVD to 30 and rFVD from 7.45 to 9.40, since a causal decoder cannot use later frames to refine a frame. With the pixel-residual stem the fully causal model reaches 27, one point from the best combination. It is also the only design that can stream: with a bidirectional encoder, prefix latents depend on frames that are not available.

\subsection{Pixel-residual stem and token budget}
\label{sec:ana_stem}

\begin{table}[t]
    \centering
    \small
    \caption{\textbf{Pixel-residual stem and token budget on K600.} Reconstruction of fully causal tokenizers with and without the stem, and of the bidirectional-decoder reference. All rows use the ablation setup; see \cref{tab:main} for the final system.}
    \label{tab:stem}
    \setlength{\tabcolsep}{4.5pt}
    \begin{tabular}{@{}l c c c c c@{}}
        \toprule
        & \multicolumn{2}{c}{\textbf{rFVD} $\downarrow$} & & & \\
        \cmidrule(lr){2-3}
        \textbf{Decoder} & $64^2$ & $224^2$ & \textbf{PSNR} $\uparrow$ & \textbf{LPIPS} $\downarrow$ & \textbf{gFVD} $\downarrow$ \\
        \midrule
        \multicolumn{6}{@{}l}{\emph{736 tokens}} \\
        Causal & 8.47 & 10.56 & 21.01 & 0.139 & 7.9 \\
        \rowcolor{cvprblue!10}Causal + stem & 5.77 & 7.47 & 24.10 & 0.103 & 5.7 \\
        Bidirectional & 3.49 & 4.62 & 22.42 & 0.110 & 5.0 \\
        \midrule
        \multicolumn{6}{@{}l}{\emph{1,216 tokens}} \\
        Causal & 5.20 & 7.72 & 21.83 & 0.120 & 4.9 \\
        \rowcolor{cvprblue!10}Causal + stem & 3.50 & 5.63 & \textbf{24.68} & \textbf{0.093} & \textbf{3.8} \\
        Bidirectional & \textbf{3.07} & \textbf{4.08} & 22.64 & 0.105 & 4.1 \\
        \bottomrule
    \end{tabular}
\end{table}

\noindent\textbf{Stem.}
We measure the effect of the stem at two token budgets (\cref{tab:stem}). The 1,216-token rows are a matched pair with the same initialization and 20-epoch schedule: the stem lowers rFVD from 5.20 to 3.50 and gFVD from 4.9 to 3.8, and the gap to the bidirectional-decoder reference at $64^2$ shrinks to 0.4 rFVD. At 736 tokens the stem row also received additional training (\cref{sec:supp_protocol}).

\noindent\textbf{Token budget.}
More derivative tokens improve reconstruction at every setting. On Kinetics-600 frame prediction, gFVD with the stem improves from 5.7 to 3.8 as the budget grows from 736 to 1,216 tokens, while on UCF-101 the two budgets are on par with the same generator (\cref{sec:supp_gen}). We use the smallest budget for long video, where the token count decides what can be trained.

\subsection{Limitations}
\label{sec:limitations}

One-minute reconstruction is tested on two datasets, one of them zero-shot. One-minute generation is shown on Ego4D only, qualitatively and without a standard distributional metric for long videos. DINOv2 is one choice of semantic backbone. We study the use of such a backbone, not which one is best, and compare with SigLIP~2 in reconstruction (\cref{sec:supp_recon}). We have not trained a text-conditioned model. Whether primitive and derivative tokens separate appearance from motion is also open: swapping them between two videos transfers appearance (\cref{sec:supp_probe}), but there is no agreed metric to test the separation.

\section{Conclusion}
\label{sec:conclusion}

\ourmodel treats video tokens the way a codec treats frames, with a full set of primitive tokens at periodic anchors and a few derivative tokens in between, read causally from a frozen semantic encoder. For a minute of video the resulting sequence is $18.8\times$ shorter than one that keeps every patch token of every frame. It keeps one-minute generation affordable and is causal enough to be generated chunk by chunk, and a standard diffusion transformer trained on it is competitive on short-clip benchmarks. Our ablations point to a broader lesson for video tokenizers: bidirectional context and pixel-level latents improve reconstruction but do not make the latents easier to generate.

\section*{LLM Usage Statement}
The authors take full responsibility for all content of this submission. LLM-based tools were used for assistance, under the authors' direction and with the authors checking the results. \emph{Writing.} An LLM assistant (Claude, Anthropic) was used to edit, rephrase and refine the authors' text and captions, to check the manuscript for consistency, and to give reviewer-style feedback. \emph{Related work and citations.} The assistant refined the authors' related-work text and checked bibliography entries against their sources. \emph{Code.} LLM-based coding assistants helped to write, refine and debug training, evaluation and plotting code. \emph{Experiments.} LLM-based tools assisted the authors in designing control experiments, in launching and monitoring training and evaluation jobs, in collecting results into tables, and in a first screening of the generated one-minute rollouts of \cref{tab:long_cap} under the authors' criterion, which the authors reviewed. \emph{Figures.} Plots and diagrams designed by the authors were drawn and refined with LLM assistance. No generative tool created or altered any frame shown in the figures; all frames are outputs of our trained models or real video. The datasets and the research ideas, claims and conclusions are the authors' own.

\clearpage
\setcounter{table}{0}
\setcounter{figure}{0}
\setcounter{equation}{0}
\renewcommand{\theHtable}{S.\arabic{table}}
\renewcommand{\theHfigure}{S.\arabic{figure}}
\renewcommand{\theHequation}{S.\arabic{equation}}
\renewcommand{\bibliographystyle}[1]{}
\renewcommand{\bibliography}[1]{}
\makeatletter\setlength{\@fptop}{0pt}\makeatother
\raggedbottom

\maketitlesupplementary
\appendix
\renewcommand{\thetable}{S\arabic{table}}
\renewcommand{\thefigure}{S\arabic{figure}}
\renewcommand{\theequation}{S\arabic{equation}}

\label{app:start}

\noindent \cref{sec:supp_protocol} gives the training and evaluation protocols for the numbers of the main paper, and \cref{sec:supp_path} traces how the final system was reached. \cref{sec:supp_method} adds method details. \cref{sec:supp_recon,sec:supp_gen,sec:supp_causality} extend the reconstruction, generation and causality results, \cref{sec:supp_long_anatomy} analyses the one-minute tokenizer over time, and \cref{sec:supp_qualitative} shows qualitative results. \cref{sec:supp_impl} lists compute and hyperparameters, \cref{sec:supp_fr4} reports the study of the tokenizer without the pretrained encoder, and \cref{sec:supp_probe} a probe that recombines the tokens of two videos. We write K600 for Kinetics-600~\cite{carreira2018short} and SSv2 for Something-Something-v2~\cite{goyal2017something}. The studies of \cref{sec:supp_long_anatomy} were run on a development prototype, as stated there.

\section{Training and Evaluation Details}
\label{sec:supp_protocol}

\paragraph{Short-clip tokenizers.}
Short-clip tokenizers start from a single-frame tokenizer trained on ImageNet~\cite{deng2009imagenet} and are trained on 16-frame K600 clips at $224^2$. Tokenizers with the pixel-residual stem are finetuned from earlier tokenizers of the same token budget. For the causality study (\cref{tab:causality}), the four tokenizers are trained with one schedule on K600 from the same single-frame initialization.

\paragraph{Generators.}
All generators are DiT~\cite{peebles2023scalable} models; their settings are listed in \cref{tab:arch_dit_supp}. DDPM~\cite{ho2020denoising} generators use $v$-prediction~\cite{salimans2022progressive} and DDIM sampling~\cite{song2021denoising}. On K600, the generator on fully causal 1,216-token latents is finetuned from an earlier generator of the same token budget.

\paragraph{Generators of the main table.}
Both K600 entries predict from the five conditioning frames alone, without class labels, and are evaluated on 50,000 clips. All UCF-101~\cite{soomro2012ucf101} and Something-Something-v2 evaluations in the main paper and in this supplement run generator and tokenizer in 32-bit floating point; the K600 generators use 16-bit transformer activations with 32-bit tokenizer decoding and 32-bit I3D features. UCF-101 rFVD draws its 50,000 clips anew in every evaluation pass.

\paragraph{One-minute models.}
The long-video tokenizer is trained with the $L_2$ loss on 1,440-frame Ego4D~\cite{grauman2022ego4d} videos at $224^2$; the causal decoder is then finetuned with the encoder frozen (\cref{tab:arch_tok_supp}). The generators on \ourmodel and on the tokenizer without the pretrained encoder in \cref{tab:long_cap,fig:long_gen,fig:supp_long_gen} are DiTs trained with the flow objective~\cite{liu2023flow,esser2024scaling} on 240-frame chunks (3,264 tokens) with Gaussian prefix noise augmentation and sampled chunk by chunk, each chunk conditioned on the last latents of the previous one.

\paragraph{Evaluation.}
FVD~\cite{unterthiner2019fvd} is computed with I3D features on 16-frame clips: rFVD between decoded clips and their inputs, gFVD between generated and real clips. Short-clip rFVD and gFVD use 50,000 clips unless stated otherwise. Clips are resized frame by frame by bilinear interpolation, without antialiasing and without cropping, when the evaluation resolution differs from the native one: $64^2$ on K600, $128^2$ on UCF-101 and Something-Something-v2, or $256^2$. The K600 protocol at $64^2$ then resizes the clip to $256^2$ and takes the central $224^2$ crop as I3D input; all other evaluations, and \cref{tab:k600_multimetric_supp}, resize to $224^2$ directly. PSNR, SSIM~\cite{wang2004image} and LPIPS~\cite{zhang2018unreasonable} are computed at the native resolution of each tokenizer, and LPIPS uses AlexNet features throughout. ImageNet rFID~\cite{heusel2017gans} of the single-frame tokenizers (\cref{tab:imagenet_stage1_supp}) is computed on the 50,000 validation images, resized and centre-cropped to $224^2$. For one-minute videos we report reconstruction PSNR, SSIM and LPIPS, the drift over the minute and the phase spread within a group, on held-out videos (\cref{tab:long_recon}), and we show generated minutes for a generator trained with the same recipe on each tokenizer (\cref{fig:long_gen,fig:supp_long_gen}).

\paragraph{Kinetics-600.}
Frame prediction follows the protocol of MAGVIT~\cite{yu2023magvit}: 5 conditioning frames and 11 predicted frames, on 50,000 clips of the validation split, which are two clips per video at linearly spaced positions times three spatial crops. A clip is 16 consecutive frames at 25 fps. FVD is computed at $64^2$ against the real clips, with one sample per clip; an evaluated clip consists of the conditioning frames as decoded by the tokenizer and the 11 generated frames. rFVD at $64^2$ uses the same 50,000 clips, and \cref{tab:k600_multimetric_supp} uses 50,000 clips of the training split.

\paragraph{Ego4D.}
The one-minute models are trained on 9,935 one-minute videos, 1,440 frames at 24 fps and $224^2$. The held-out videos come from recordings that share no participant with the training videos. For reconstruction, PSNR, SSIM and LPIPS (AlexNet features) are averaged over videos for every frame and then over frames (\cref{tab:long_recon}). Continuation is evaluated on 455 distinct held-out videos. Each rollout starts from the first 40 ground-truth frames (1.67 seconds) and is generated in seven calls with one seed, the latents being handed from one call to the next as described in \cref{sec:supp_method}. A rollout counts as coherent if its scene does not break down into distorted textures, and a rollout whose picture fades to dark without such a breakdown also counts as coherent. Every rollout was rendered as a sheet of frames over the minute; an LLM-based tool made a first assignment under this criterion, which the authors reviewed.

\paragraph{YouCook2.}
The zero-shot evaluation uses the first 256 videos of the 456-video validation split~\cite{zhou2018towards} in its stored order, without selection by content. From each video the central 1,440 consecutive frames at the native 25 fps are taken, centre-cropped to $224^2$. The tokenizer is the one trained on Ego4D, without finetuning, and the reconstruction metrics are those used on Ego4D.

\paragraph{Something-Something-v2.}
A clip is 16 frames sampled uniformly over the whole video, centre-cropped to $224^2$. Generated clips are evaluated at $128^2$ and $224^2$ against the 24,777-video validation split, in four sets of 10,000 generated clips, each against 10,000 validation clips with one clip per video.

\paragraph{Sample counts on UCF-101.}
UCF-101 results use the 9,537 training videos of Split~1. rFVD is computed on 50,000 clips of 16 consecutive frames drawn from them. The generators are trained on 16-frame clips of these videos, and the class labels of the generated samples follow their class distribution. UCF-101 gFVD depends strongly on the number of generated samples. \cref{tab:main} uses 10,000 samples, the protocol of MAGVIT~\cite{yu2023magvit}, MAGVIT-v2~\cite{yu2024magvitv2}, LARP~\cite{wang2025larp}, AdapTok~\cite{li2025adaptok}, DeRA~\cite{guo2025dera} and W.A.L.T.~\cite{gupta2024walt}. TATS~\cite{ge2022long} uses 2,048 samples, and OmniTokenizer~\cite{wang2024omnitokenizer}, SweetTok~\cite{tan2025sweettok}, VFRTok~\cite{zhong2025vfrtok} and ViTok~\cite{hansen2025vitok} do not state a count. The table lists W.A.L.T.-L, 46.0; W.A.L.T.-XL reaches 36.0. \ourmodel operates at $224^2$. For the $256^2$ block of \cref{tab:main} its outputs are resized to $256^2$ by bilinear interpolation before evaluation; the tables of this supplement that say ``native'' evaluate at $224^2$. \cref{tab:videorae} repeats the evaluation with 2,048 samples at native resolution for VideoRAE, which reports gFVD with continuous latents and a diffusion transformer.

\begin{table}[t]
    \centering
    \small
    \caption{\textbf{UCF-101 class-conditional generation with continuous latents and a diffusion transformer, 2K generated samples.} Rows above the double rule are from Table~5 of VideoRAE~\cite{xie2026videorae}. ``Latent'' is the number of tokens $\times$ the channel width.}
    \label{tab:videorae}
    \setlength{\tabcolsep}{3pt}
    \begin{tabular}{l c c c}
        \toprule
        \textbf{Method} & \textbf{Res.} & \textbf{Latent} & \textbf{gFVD} $\downarrow$ \\
        \midrule
        Latte & $256^2$ & $4096\times4$ & 478 \\
        LeanVAE & $256^2$ & $4096\times4$ & 164 \\
        LTX-VAE & $256^2$ & $128\times128$ & 161 \\
        VideoRAE (VideoMAEv2) & $224^2$ & $512\times32$ & 99 \\
        VideoRAE (V-JEPA~2) & $256^2$ & $512\times32$ & 93 \\
        \midrule
        WF-VAE & $256^2$ & $4096\times16$ & 371 \\
        LeanVAE & $256^2$ & $4096\times16$ & 175 \\
        DeCo-VAE & $256^2$ & $4096\times16$ & 166 \\
        CogVideoX-VAE & $256^2$ & $4096\times16$ & 151 \\
        Wan2.1-VAE & $256^2$ & $4096\times16$ & 126 \\
        VideoRAE (VideoMAEv2) & $224^2$ & $1024\times64$ & 103 \\
        VideoRAE (V-JEPA~2) & $256^2$ & $1024\times64$ & 94 \\
        \midrule\midrule
        \rowcolor{cvprblue!10}\ourmodel\ (Ours) & $224^2$ & $736\times64$ & 82 \\
        \bottomrule
    \end{tabular}
\end{table}

\section{Development Path}
\label{sec:supp_path}

This section summarizes the design steps behind the final system. On K600 with 1,216 tokens, making the decoder causal raises rFVD at $64^2$ from 3.07 to 5.20, and the pixel-residual stem recovers the loss, with rFVD 3.50 (\cref{tab:stem}); longer training lowers it to 2.50 (\cref{tab:k600_capacity_supp}).

\section{Additional Method Details}
\label{sec:supp_method}

\noindent\textbf{Scope of prefix invariance and the link between chunks.}
\cref{eq:prefix_invariance} holds for whole groups inside one chunk. For 16-frame models a group is one frame. For one-minute models a group holds four frames that are pooled together, so a prefix ends at a group boundary and the last group is complete before it is encoded. A generator call covers one chunk of 240 frames, that is 60 groups with 6 anchors, or $6\cdot 256+54\cdot 32=3{,}264$ tokens. The last 10 groups of a chunk, 40 frames and $256+9\cdot 32=544$ tokens, are copied to the first 10 positions of the next chunk. Group 50 of a chunk is an anchor, so anchors stay anchors and no token changes its role. In the new chunk these latents receive the positions of groups 0 to 9, the clean timestep and the prefix mask. They were generated to match latents that the tokenizer forms with the 50 earlier groups of the old chunk as context, whereas the prefix of a training chunk is encoded without that context. This shift is outside \cref{eq:prefix_invariance}. The generator is trained with Gaussian noise on the prefix latents to tolerate it, and \cref{fig:long_gen,fig:supp_long_gen,fig:supp_long_gen_full} test it empirically. A rollout from a 40-frame prefix has seven calls and $40+7\cdot 200=1{,}440$ frames. Each call after the first keeps its 2,720 new tokens, so the video has $3{,}264+6\cdot 2{,}720=19{,}584$ distinct tokens, while the seven calls process $7\cdot 3{,}264=22{,}848$ token positions. The result is decoded in six chunks of 240 frames without overlap.

\noindent\textbf{Chunk-wise generation.}
To generate 1,440 frames, the first chunk of 240 frames is sampled unconditionally or from a ground-truth prefix of 40 frames. Each later chunk is conditioned on the last latents of the previous one, which are projected back to the unit sphere, without decoding to pixels in between. The 20-second continuations start from a ground-truth prefix of 480 frames, that is two chunks. Gaussian noise on the prefix during training makes the model tolerate generated prefixes.

\section{Additional Reconstruction Results}
\label{sec:supp_recon}

\begin{figure*}[t]
    \centering
    \includegraphics[width=\textwidth]{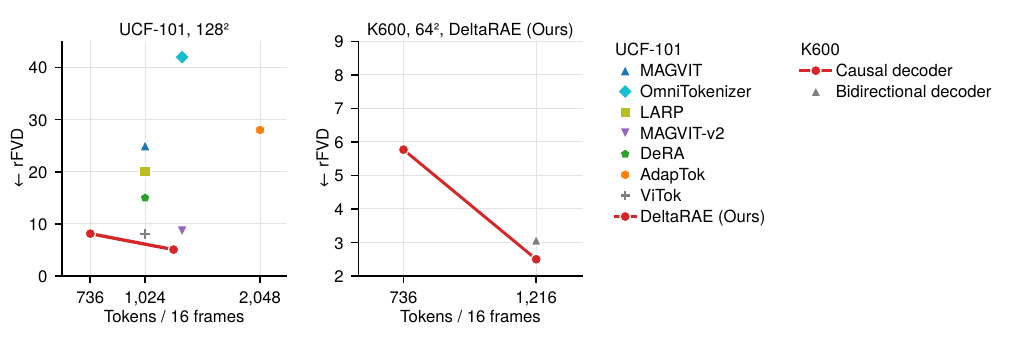}
    \caption{\textbf{Reconstruction quality against token count} on UCF-101 (left, with prior tokenizers of \cref{tab:main}) and on K600 (right).}
    \label{fig:pareto_rfvd}
\end{figure*}

\paragraph{Frozen encoders.}
\cref{tab:k600_multimetric_supp} reports K600 reconstruction with two frozen encoders, DINOv2~\cite{oquab2023dinov2} ViT-B/14 at $224^2$ and SigLIP~2~\cite{tschannen2025siglip} ViT-B/16 at $256^2$, on 50,000 clips of the training split. With DINOv2 every metric improves as the token budget grows from 736 to 1,216 tokens: rFVD at $128^2$ falls from 3.75 to 3.21, PSNR rises from 22.56 to 22.81 dB and LPIPS falls from 0.107 to 0.102. Both are frozen semantic encoders and both can serve as the backbone of the tokenizer. All other experiments use DINOv2.

\begin{table*}[t]
    \centering
    \small
    \caption{\textbf{K600 reconstruction with two frozen encoders,} evaluated with the same protocol.}
    \label{tab:k600_multimetric_supp}
    \setlength{\tabcolsep}{4.5pt}
\begin{tabular}{l c | c c c c c c}
        \toprule
        \textbf{Frozen encoder} & \textbf{Tokens / 16f} & \textbf{rFVD native} $\downarrow$ & \textbf{rFVD $128^2$} $\downarrow$ & \textbf{rFID} $\downarrow$ & \textbf{PSNR (dB)} $\uparrow$ & \textbf{SSIM} $\uparrow$ & \textbf{LPIPS} $\downarrow$ \\
        \midrule
        DINOv2 ViT-B/14 ($224^2$) & 736 & 4.31 & 3.75 & 0.99 & 22.56 & 0.708 & 0.107 \\
        DINOv2 ViT-B/14 ($224^2$) & 1,216 & 3.77 & 3.21 & 0.92 & 22.81 & 0.718 & 0.102 \\
        \midrule
        SigLIP~2 ViT-B/16 ($256^2$) & 736 & 9.59 & 9.09 & 1.13 & 21.74 & 0.675 & 0.137 \\
        SigLIP~2 ViT-B/16 ($256^2$) & 1,216 & 10.05 & 9.54 & 1.12 & 22.08 & 0.687 & 0.129 \\
        SigLIP~2 ViT-B/16 ($256^2$, dense) & 4,096 & 8.08 & 7.62 & 1.05 & 22.91 & 0.715 & 0.110 \\
        \bottomrule
    \end{tabular}%
\end{table*}

\paragraph{Token budget and decoder causality on K600.}
K600 reconstruction is also measured on the 50,000 validation clips of the frame-prediction benchmark, with 2 clips $\times$ 3 crops per video (\cref{tab:k600_capacity_supp}). More tokens lower rFVD at $64^2$ from 5.77 to 3.50 for the fully causal tokenizer. With longer training it reaches 2.50 at 1,216 tokens, with 26.20 dB PSNR, below the 3.07 of the tokenizer with a bidirectional decoder (\cref{tab:causality}). The effect of the stem is analysed in \cref{tab:stem}.

\begin{table*}[t]
    \centering
    \small
    \caption{\textbf{K600 reconstruction by tokenizer and token budget.} The native resolution is $224^2$, and $256^2$ without the pretrained encoder (\cref{sec:supp_fr4}).}
    \label{tab:k600_capacity_supp}
    \setlength{\tabcolsep}{8pt}
    \begin{tabular}{l c | c c c c}
        \toprule
        \textbf{Tokenizer} & \textbf{Tokens / 16f} & \textbf{rFVD $64^2$} $\downarrow$ & \textbf{rFVD native} $\downarrow$ & \textbf{PSNR (dB)} $\uparrow$ & \textbf{LPIPS} $\downarrow$ \\
        \midrule
        \rowcolor{cvprblue!14}\ourmodel & 736 & 5.77 & 7.47 & 24.10 & 0.103 \\
        \rowcolor{cvprblue!14}\ourmodel & 1,216 & 3.50 & 5.63 & 24.68 & 0.093 \\
        \rowcolor{cvprblue!14}\ourmodel, longer training & 1,216 & 2.50 & 4.24 & 26.20 & 0.077 \\
        \midrule
        w/o pretrained encoder & 736 & 3.44 & 4.61 & 30.01 & 0.060 \\
        w/o pretrained encoder & 1,216 & 3.09 & 4.98 & 30.81 & 0.053 \\
        \bottomrule
    \end{tabular}
\end{table*}

\paragraph{Image pretraining.}
\cref{tab:imagenet_stage1_supp} reports the single-frame DINOv2 tokenizers trained on ImageNet-1K from which the video tokenizers are initialized. With a latent width of 64 they reach 0.75 rFID and 22.97 dB PSNR.

\begin{table}[t]
    \centering
    \small
    \caption{\textbf{Single-frame DINOv2 tokenizers on ImageNet-1K,} 256 tokens per image.}
    \label{tab:imagenet_stage1_supp}
    \setlength{\tabcolsep}{6pt}
    \begin{tabular}{c | c c c}
        \toprule
        \textbf{Latent width} & \textbf{rFID} $\downarrow$ & \textbf{PSNR} $\uparrow$ & \textbf{SSIM} $\uparrow$ \\
        \midrule
        16 & 1.94 & 20.69 & 0.570 \\
        32 & 0.96 & 22.02 & 0.630 \\
        \rowcolor{cvprblue!12}64 & 0.75 & 22.97 & 0.673 \\
        \bottomrule
    \end{tabular}
\end{table}

\section{Additional Generation Results}
\label{sec:supp_gen}

\paragraph{UCF-101, Something-Something-v2 and K600.}
\cref{tab:ucf_gen_supp,tab:ssv2_supp,tab:k600_gen_supp} list generation results on the three 16-frame benchmarks.
\begin{enumerate}[leftmargin=*, noitemsep, topsep=2pt]
    \item \emph{UCF-101.} With the same generator the two token budgets are on par: 45 and 44 in the ablation setup of \cref{tab:ucf_gen_supp}, and 27 and 26 with the flow generator (\cref{tab:causality}).
    \item \emph{Something-Something-v2.} Horizontal flips are disabled in training so that left and right keep their meaning. Both token budgets reach 44 gFVD at $128^2$ (\cref{tab:ssv2_supp}).
    \item \emph{K600 frame prediction.} \cref{tab:k600_gen_supp} gives the final system at $64^2$ and at $224^2$, 2.9 and 4.6, together with the DDPM comparison of \cref{sec:supp_fr4}.
\end{enumerate}

\begin{table*}[t]
    \centering
    \small
    \caption{\textbf{UCF-101 class-conditional generation at different token counts,} with the same DDPM generator recipe and the same evaluation in all rows; 4,096 tokens are the dense grid. These numbers are from the ablation setup, not from the final system (\cref{tab:main}).}
    \label{tab:ucf_gen_supp}
    \setlength{\tabcolsep}{8pt}
    \begin{tabular}{c c | c c}
        \toprule
        \textbf{Pretrained encoder} & \textbf{Tokens / 16f} & \textbf{gFVD $128^2$} $\downarrow$ & \textbf{gFVD $224^2$} $\downarrow$ \\
        \midrule
        \checkmark & 736 & 45 & 49 \\
        \checkmark & 1,216 & 44 & 47 \\
        \checkmark & 4,096 & 38 & 41 \\
        $\times$ & 736 & 226 & --- \\
        \bottomrule
    \end{tabular}
\end{table*}

\begin{table*}[t]
    \centering
    \small
    \caption{\textbf{Something-Something-v2 class-conditional generation with 174 classes.} DDPM generator, 10,000 generated clips against the validation split (\cref{sec:supp_protocol}). Both rows share the generator recipe and the evaluation.}
    \label{tab:ssv2_supp}
    \setlength{\tabcolsep}{8pt}
    \begin{tabular}{c | c c}
        \toprule
        \textbf{Tokens / 16f} & \textbf{gFVD $128^2$} $\downarrow$ & \textbf{gFVD $224^2$} $\downarrow$ \\
        \midrule
        736 & 44 & 50 \\
        1,216 & 44 & 52 \\
        \bottomrule
    \end{tabular}%
\end{table*}

\begin{table*}[t]
    \centering
    \small
    \caption{\textbf{K600 frame prediction.} The first row is the final system of \cref{tab:main}; the other rows are from the ablation setup and share one DDPM generator recipe.}
    \label{tab:k600_gen_supp}
    \setlength{\tabcolsep}{8pt}
    \begin{tabular}{l c c | c c}
        \toprule
        \textbf{Tokenizer} & \textbf{Generator} & \textbf{Tokens / 16f} & \textbf{gFVD $64^2$} $\downarrow$ & \textbf{gFVD $224^2$} $\downarrow$ \\
        \midrule
        \rowcolor{cvprblue!14}\ourmodel, final system & flow & 1,216 & 2.9 & 4.6 \\
        \midrule
        Frozen DINOv2 & DDPM & 1,216 & 4.3 & 5.7 \\
        Frozen DINOv2 & DDPM & 736 & 5.0 & 6.1 \\
        w/o pretrained encoder & DDPM & 1,216 & 19.7 & --- \\
        w/o pretrained encoder & DDPM & 736 & 22.7 & --- \\
        \bottomrule
    \end{tabular}
\end{table*}

\section{Encoder and Decoder Causality}
\label{sec:supp_causality}

\begin{table*}[t]
    \centering
    \small
    \caption{\textbf{Encoder and decoder causality.} The four 736-token combinations share one training schedule and one flow generator recipe. These numbers are from the ablation setup, not from the final system (\cref{tab:main}).}
    \label{tab:causality}
    \setlength{\tabcolsep}{4pt}
\begin{tabular}{l l c | c c | c c | c c | c c | c c}
    \toprule
        & & & \multicolumn{2}{c|}{\textbf{Prefix-invariant}} & \multicolumn{2}{c|}{\textbf{K600 rFVD} $\downarrow$} & \multicolumn{2}{c|}{\textbf{K600 gFVD} $\downarrow$} & \multicolumn{2}{c|}{\textbf{UCF rFVD} $\downarrow$} & \multicolumn{2}{c}{\textbf{UCF gFVD} $\downarrow$} \\
        \textbf{Encoder} & \textbf{Decoder} & \textbf{Tokens} & encoder & decoder & $64^2$ & $224^2$ & $64^2$ & $224^2$ & $128^2$ & $224^2$ & $128^2$ & $224^2$ \\
    \midrule
        Bidirectional & Bidirectional & 736 & $\times$ & $\times$ & 5.35 & 6.50 & --- & --- & 6.84 & 7.58 & 31 & 35 \\
        Causal & Bidirectional & 736 & \checkmark & $\times$ & 6.05 & 7.28 & 6.5 & --- & 7.45 & 8.27 & 26 & 28 \\
        Bidirectional & Causal & 736 & $\times$ & \checkmark & 6.36 & 7.12 & --- & --- & 8.27 & 9.14 & 28 & 30 \\
        Causal & Causal & 736 & \checkmark & \checkmark & 8.47 & 10.56 & 7.9 & 9.5 & 9.40 & 10.47 & 30 & 32 \\
        \rowcolor{cvprblue!10}Causal + stem & Causal & 736 & \checkmark & \checkmark & 5.77 & 7.47 & 5.7 & --- & 8.38 & 9.38 & 27 & 29 \\
    \midrule
        Causal & Bidirectional & 1,216 & \checkmark & $\times$ & 3.07 & 4.08 & 4.1 & 5.1 & 4.68 & 5.49 & 26 & 29 \\
        \rowcolor{cvprblue!10}Causal + stem & Causal & 1,216 & \checkmark & \checkmark & 3.50 & 5.63 & 3.8 & 5.5 & 5.20 & 5.99 & 26 & 28 \\
    \bottomrule
\end{tabular}%
\end{table*}

\paragraph{All four combinations.}
\cref{tab:causality} lists the four combinations with reconstruction and generation on both datasets; the numerical causality test of \cref{sec:asymmetric_causality} is the first item below.
\begin{enumerate}[leftmargin=*, noitemsep, topsep=2pt]
    \item \emph{Perturbation test.} We replace the frames after a prefix of 4, 8 or 12 frames by uniform noise. With a causal encoder the prefix latents change by exactly zero. The decoder is tested on its own: with the prefix latents held fixed and the later latents replaced, the prefix decoded by a causal decoder changes by exactly zero. From pixels to pixels the prefix is unchanged only when both encoder and decoder are causal. With bidirectional attention the reconstructed prefix changes by up to 0.13, 0.077 and 0.023 for the three prefix lengths.
    \item \emph{Bidirectional encoder and decoder.} This tokenizer has the lowest UCF-101 rFVD, 6.84, and the highest gFVD of the four combinations, 31.
    \item \emph{Causal decoder and stem.} With a causal encoder, making the decoder causal as well raises UCF-101 gFVD from 26 to 30, and the stem brings it back to 27.
\end{enumerate}

\begin{table*}[t]
    \centering
    \small
    \caption{\textbf{Tokenizer architecture and training settings} of \ourmodel and of the tokenizer without the pretrained encoder (\cref{sec:supp_fr4}).}
    \label{tab:arch_tok_supp}
    \setlength{\tabcolsep}{4pt}
    \newcommand{\both}[1]{\multicolumn{2}{>{\centering\arraybackslash}p{\dimexpr\linewidth-3.2cm-4\tabcolsep-2\arrayrulewidth\relax}}{#1}}
    \begin{tabularx}{\linewidth}{>{\raggedright\arraybackslash}p{3.2cm} | C C}
        \toprule
        \textbf{Setting} & \textbf{\ourmodel\ (frozen DINOv2)} & \textbf{w/o pretrained encoder} \\
        \midrule
        Input resolution & $224^2$ & $256^2$ \\
        Encoder input & DINOv2 ViT-B/14 features, with the pixel-residual stem & pixel patches \\
        Encoder layers & 4 & 12 \\
        \midrule
        Decoder layers & \both{12} \\
        Width, heads & \both{$D=768$, 12 heads} \\
        Latent width, normalization & \both{64, unit sphere} \\
        Position encoding & \both{2D sine-cosine, added (space); 1D RoPE~\cite{su2021roformer} over the frame or group index (time)} \\
        Primitive and derivative tokens & \both{$N_a = 256$; $N_d \in \{32, 64\}$ at 16 frames, $N_d = 32$ at one minute} \\
        Anchors and chunks & \both{16 frames: one anchor, one chunk; one minute: an anchor every 10 groups, chunks of 60 groups} \\
        Attention masks & \both{causal encoder, block-causal across chunks at one minute; causal decoder} \\
        Training at one minute & \both{$L_2$, then decoder finetuning with the encoder frozen} \\
        Optimizer, peak learning rate & \both{AdamW, $10^{-4}$} \\
        \bottomrule
    \end{tabularx}
\end{table*}

\begin{table*}[t]
    \centering
    \small
    \caption{\textbf{Generator settings} of the entries of \cref{tab:main} on UCF-101 and K600, by token budget, and of the SSv2 and one-minute models.}
    \label{tab:arch_dit_supp}
    \setlength{\tabcolsep}{4pt}
    \begin{tabularx}{\linewidth}{>{\raggedright\arraybackslash}p{3.4cm} | C C | C C | C | C}
        \toprule
        \textbf{Setting} & \multicolumn{2}{c|}{\textbf{UCF-101}} & \multicolumn{2}{c|}{\textbf{K600}} & \multirow{2}{*}{\textbf{SSv2}} & \multirow{2}{*}{\textbf{Ego4D}} \\
        Tokens per 16 frames & 736 & 1,216 & 736 & 1,216 &  &  \\
        \midrule
        Objective & flow & flow & flow & flow & DDPM & flow \\
        Layers & 28 & 28 & 28 & 28 & 28 & 28 \\
        Width & 1,152 & 1,152 & 1,152 & 1,152 & 1,152 & 1,152 \\
        Class labels & yes & yes & no & no & yes & no \\
        \midrule
        Optimizer & AdamW & AdamW & AdamW & AdamW & AdamW & AdamW \\
        Learning rate & $10^{-4}$ & $10^{-4}$ & $3\times10^{-4}$ & $3.5\times10^{-4}$ & $2\times10^{-4}$ & $2\times10^{-4}$ \\
        Beta1 & 0.9 & 0.9 & 0.9 & 0.9 & 0.9 & 0.9 \\
        Beta2 & 0.99 & 0.99 & 0.99 & 0.99 & 0.99 & 0.99 \\
        Weight decay & 0.01 & 0.01 & 0.01 & 0.01 & 0.01 & 0.01 \\
        Scheduler type & cosine & cosine & cosine & cosine & cosine & cosine \\
        Warmup epochs & 5 & 5 & 1 & 1 & 5 & 26 \\
        Batch size & 256 & 256 & 1,024 & 1,024 & 256 & 128 \\
        Epochs & 12,000 & 1,500 & 200 & 400 & 85 & 640 \\
        \bottomrule
    \end{tabularx}
\end{table*}

\section{One-Minute Tokenizer: Design and Temporal Analysis}
\label{sec:supp_long_anatomy}

The studies of this section were run during development on a prototype of the one-minute tokenizer. It has the token schedule, the masks and the training stages of the final model, but a trainable pixel patch trunk at $256^2$ in place of the frozen DINOv2 encoder, which is the architecture of the tokenizer without the pretrained encoder in \cref{tab:long_recon}. Its numbers are those of the prototype. The final one-minute \ourmodel is reported in \cref{tab:long_recon,fig:supp_per_frame_curves}.

\paragraph{Target tokens of the decoder.}
\cref{tab:long_design_ablation_supp} compares two choices for the prototype. A derivative group has 32 latent tokens. The decoder can use 32 target tokens for such a group, or keep the full grid of 256 target tokens that attend to the 32 latents. At the same latent budget and the same training step, the full grid lowers pixel MSE by 33.3\%, from 0.0258 to 0.0172, and raises PSNR from 15.88 to 17.64 dB without adding latent tokens. These two runs use 12,864 latent tokens per minute, fewer than the 19,584 of the final schedule.

\begin{table*}[t]
    \centering
    \small
    \caption{\textbf{Target tokens per derivative group in the decoder of the one-minute prototype on Ego4D,} with the same latent tokens and the same training step. Anchor groups have 256 target tokens in both rows.}
    \label{tab:long_design_ablation_supp}
    \setlength{\tabcolsep}{6pt}
    \begin{tabular}{c c c | c c}
        \toprule
        \makecell[b]{\textbf{Target tokens per}\\\textbf{derivative group}} & \makecell[b]{\textbf{Latent tokens}\\\textbf{/ min}} & \textbf{Step} & \makecell[b]{\textbf{Pixel}\\\textbf{MSE} $\downarrow$} & \makecell[b]{\textbf{PSNR}\\\textbf{(dB)} $\uparrow$} \\
        \midrule
        32 & 12,864 & 5,400 & 0.02580 & 15.88 \\
        256 & 12,864 & 5,400 & 0.01720 & 17.64 \\
        \bottomrule
    \end{tabular}
\end{table*}

\paragraph{Stability within an anchor period and across chunks.}
\cref{tab:long_anatomy_supp} analyses the reconstruction of the prototype over time on 256 held-out Ego4D videos and on 256 YouCook2 videos, a dataset that was not used in training; \cref{tab:long_deflicker_supp} gives its overall metrics on both.
\begin{enumerate}[leftmargin=*, noitemsep, topsep=2pt]
    \item \emph{Within an anchor period.} Anchor groups reconstruct at 26.27 dB on Ego4D and 24.54 dB on YouCook2. The nine derivative groups that follow stay flat, at 24.85 dB for the first and 24.73 dB for the last on Ego4D and at 23.69 and 23.58 dB on YouCook2, so the error does not accumulate between anchors.
    \item \emph{Across chunks.} Over the six 10-second chunks of a minute, LPIPS on Ego4D changes by $-0.004$ and PSNR on YouCook2 moves from 23.74 to 23.50 dB, a drift of $-0.30$ dB per minute. The temporal difference error at chunk boundaries, 0.064, is close to its value at ordinary group boundaries, 0.060.
\end{enumerate}

\begin{table*}[t]
    \centering
    \small
    \caption{\textbf{One-minute reconstruction of the prototype by group type} on held-out Ego4D and on YouCook2.}
    \label{tab:long_anatomy_supp}
    \setlength{\tabcolsep}{4pt}
    \begin{tabular}{l | c c | c c}
        \toprule
        & \multicolumn{2}{c|}{\textbf{Ego4D, held-out}} & \multicolumn{2}{c}{\textbf{YouCook2, unseen}} \\
        \textbf{Slice} & \textbf{PSNR (dB)} $\uparrow$ & \textbf{LPIPS} $\downarrow$ & \textbf{PSNR (dB)} $\uparrow$ & \textbf{LPIPS} $\downarrow$ \\
        \midrule
        Anchor groups (256 tokens) & 26.27 & 0.141 & 24.54 & 0.154 \\
        Derivative groups (32 tokens) & 24.67 & 0.168 & 23.52 & 0.187 \\
        \quad first derivative group after an anchor & 24.85 & --- & 23.69 & --- \\
        \quad last derivative group before the next anchor & 24.73 & --- & 23.58 & --- \\
        \bottomrule
    \end{tabular}
\end{table*}

\begin{table*}[t]
    \centering
    \small
    \caption{\textbf{The one-minute prototype on held-out Ego4D and on YouCook2.}}
    \label{tab:long_deflicker_supp}
    \setlength{\tabcolsep}{6pt}
    \begin{tabular}{l | c c c}
        \toprule
        \textbf{Data} & \makecell[b]{\textbf{PSNR}\\\textbf{(dB)} $\uparrow$} & \textbf{SSIM} $\uparrow$ & \textbf{LPIPS} $\downarrow$ \\
        \midrule
        Ego4D, held-out & 24.83 & 0.693 & 0.166 \\
        YouCook2, not used in training & 23.62 & 0.728 & 0.183 \\
        \bottomrule
    \end{tabular}
\end{table*}

\section{Qualitative Results}
\label{sec:supp_qualitative}

\begin{figure*}[t]
    \centering
    \includegraphics[width=\linewidth]{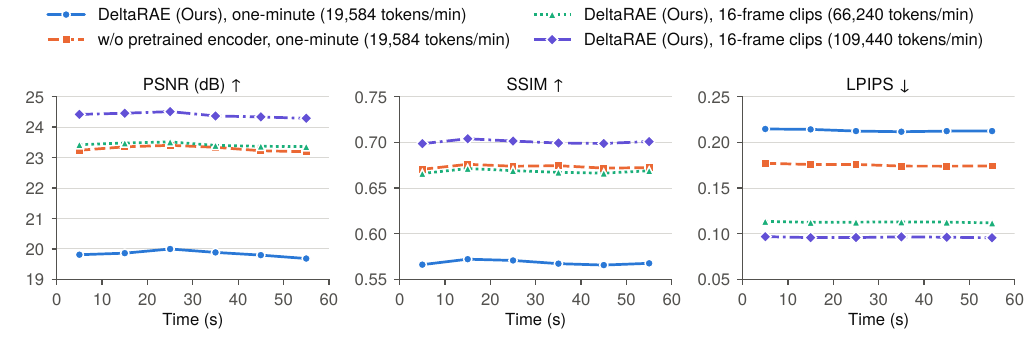}
    \caption{\textbf{One-minute reconstruction quality over time} for tokenizers of \cref{tab:long_recon}, averaged over each 10-second segment.}
    \label{fig:supp_per_frame_curves}
\end{figure*}

\begin{figure*}[t]
    \centering
    {\sffamily\let\small\footnotesize\input{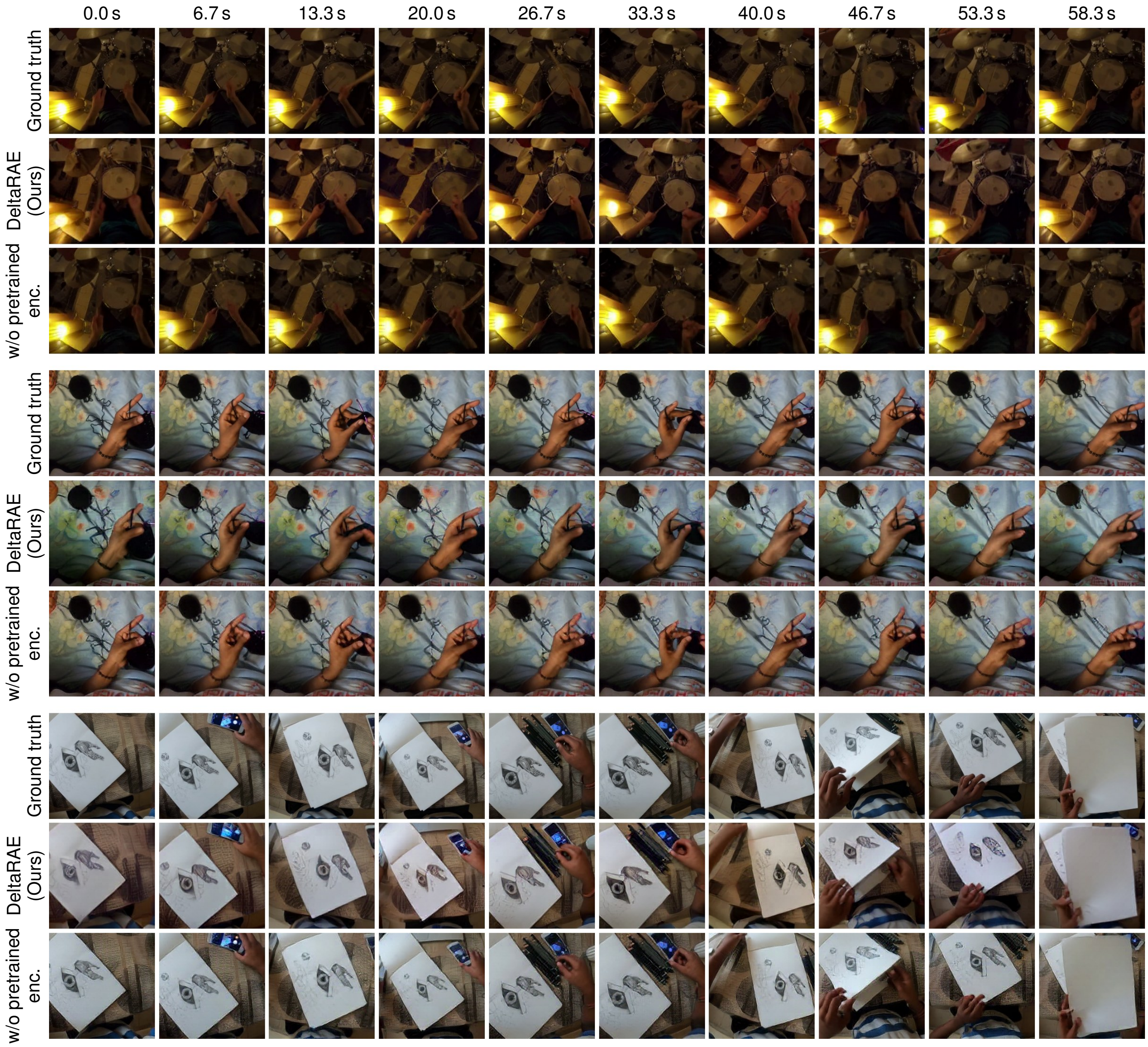}}
    \caption{\textbf{One-minute reconstruction on three held-out Ego4D videos,} with 19,584 tokens per minute for both tokenizers.}
    \label{fig:supp_long_recon}
\end{figure*}

\begin{figure*}[t]
    \centering
    {\sffamily\let\small\footnotesize\input{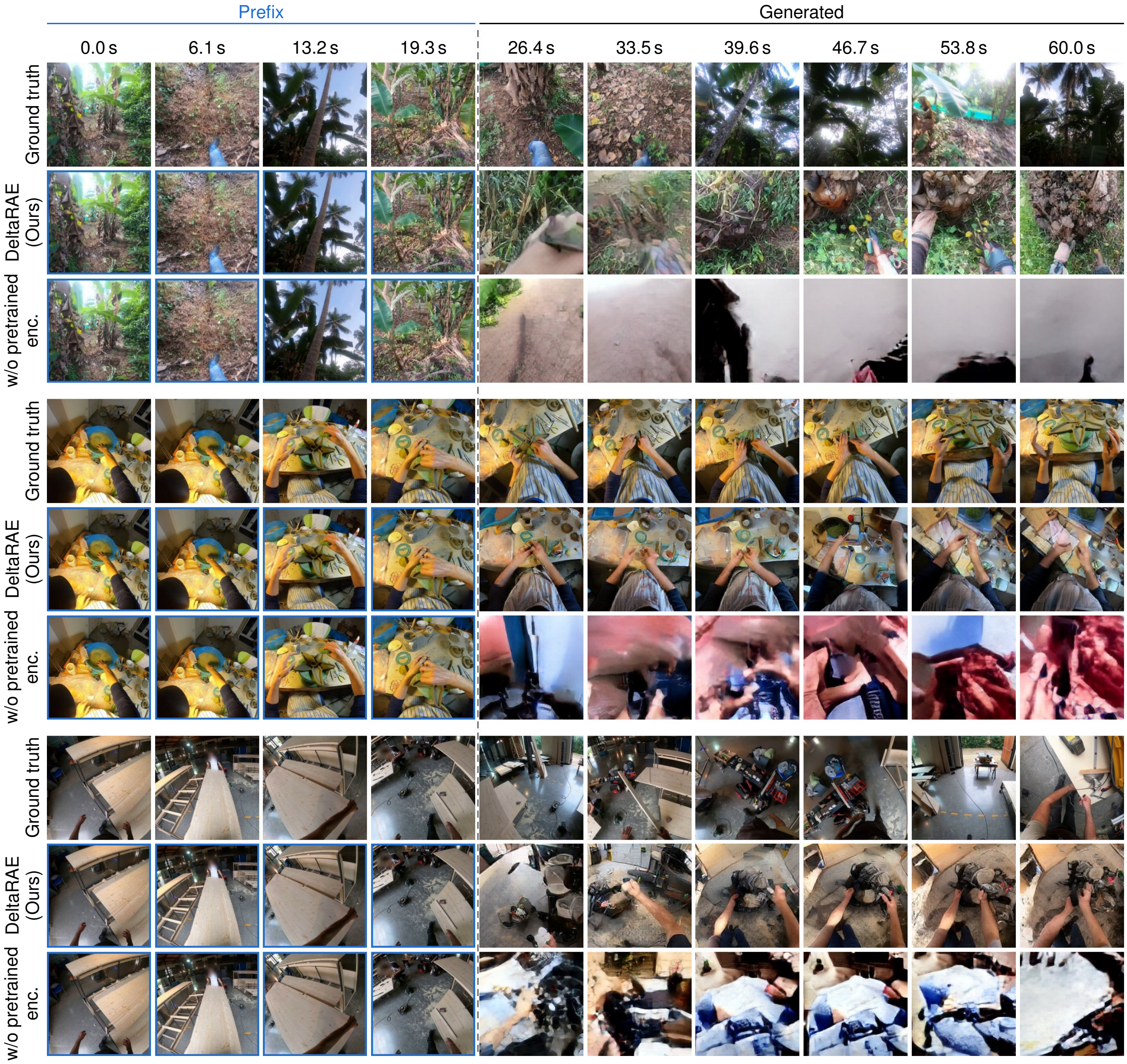}}
    \caption{\textbf{One-minute continuation from a 20-second prefix on three held-out Ego4D videos,} with the same generator recipe for both tokenizers. Framed frames lie in the prefix.}
    \label{fig:supp_long_gen}
\end{figure*}

\begin{figure*}[t]
    \centering
    {\sffamily\let\small\footnotesize\input{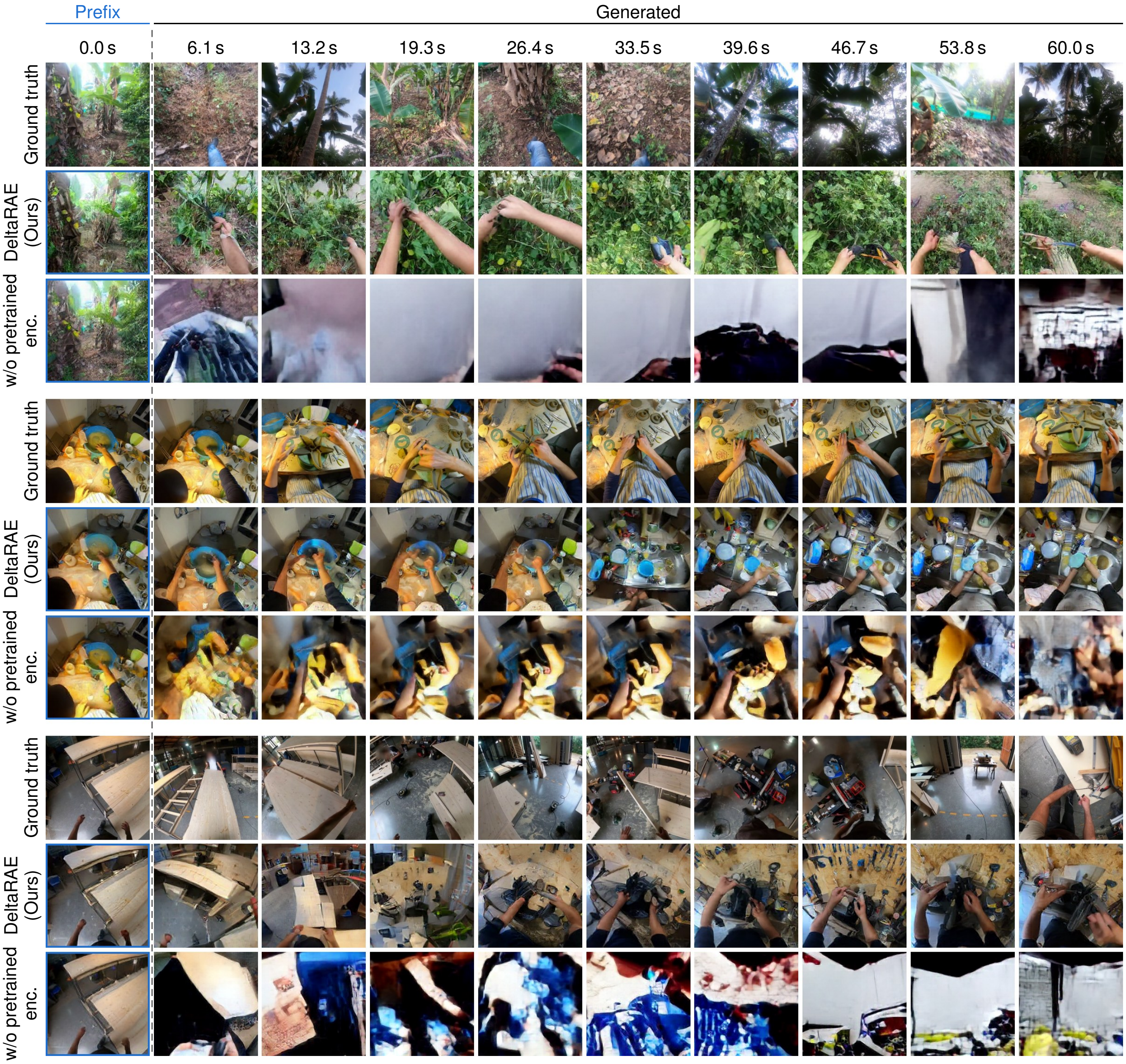}}
    \caption{\textbf{One-minute continuation from a 1.67-second prefix on three held-out Ego4D videos,} with the same generator recipe for both tokenizers. Only the first frame shown lies in the prefix.}
    \label{fig:supp_long_gen_full}
\end{figure*}

\begin{figure*}[t]
    \centering
    \includegraphics[width=0.98\linewidth]{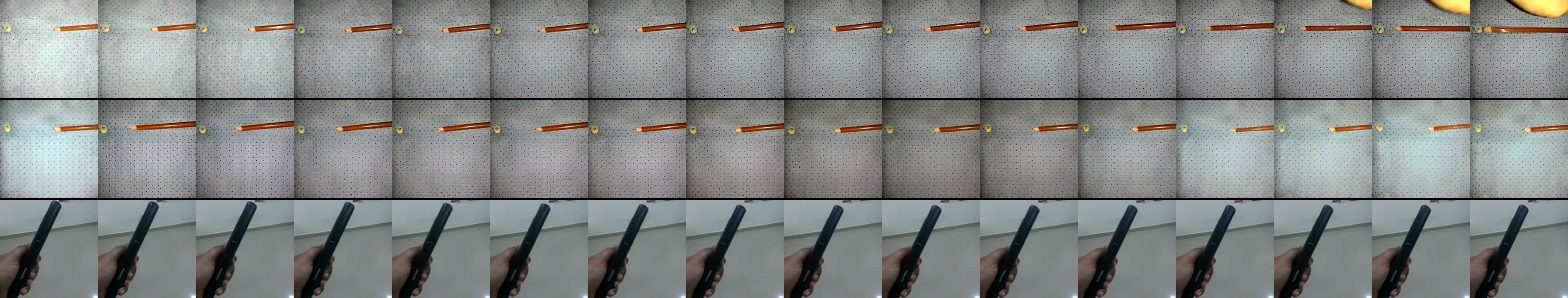}\\[3pt]
    \includegraphics[width=0.98\linewidth]{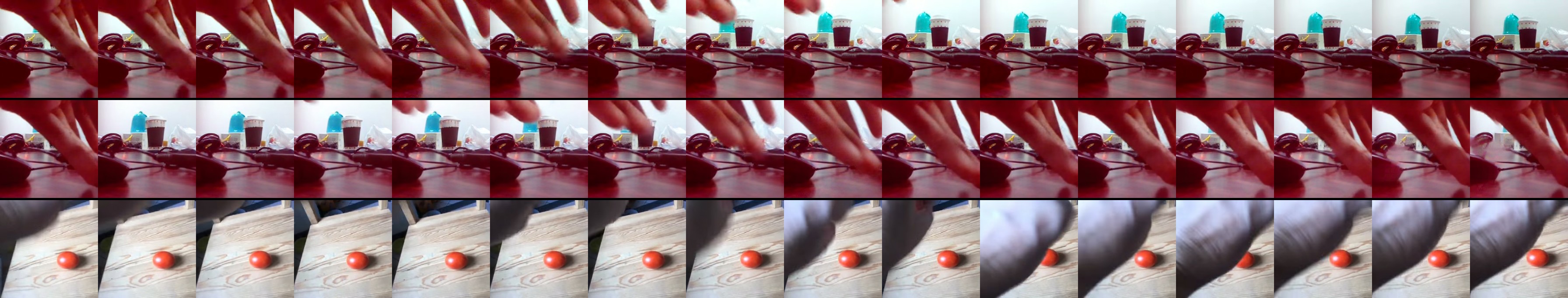}\\[3pt]
    \includegraphics[width=0.98\linewidth]{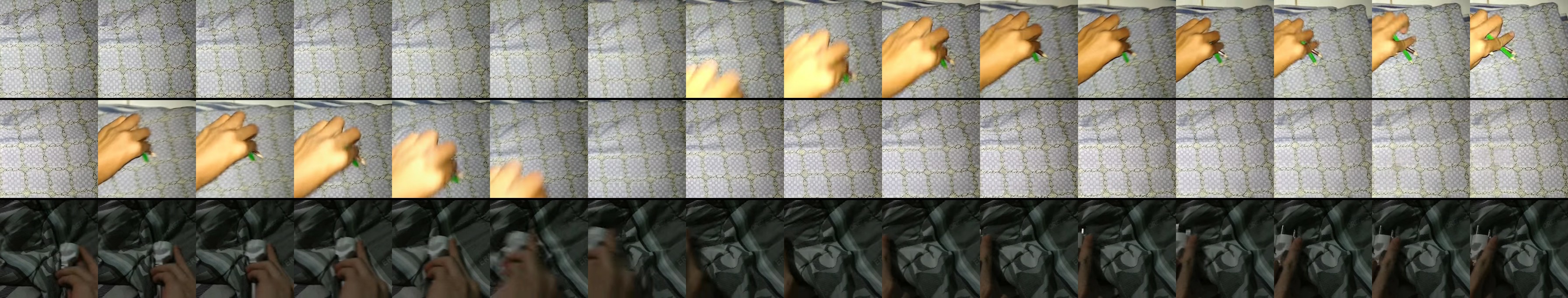}\\[3pt]
    \includegraphics[width=0.98\linewidth]{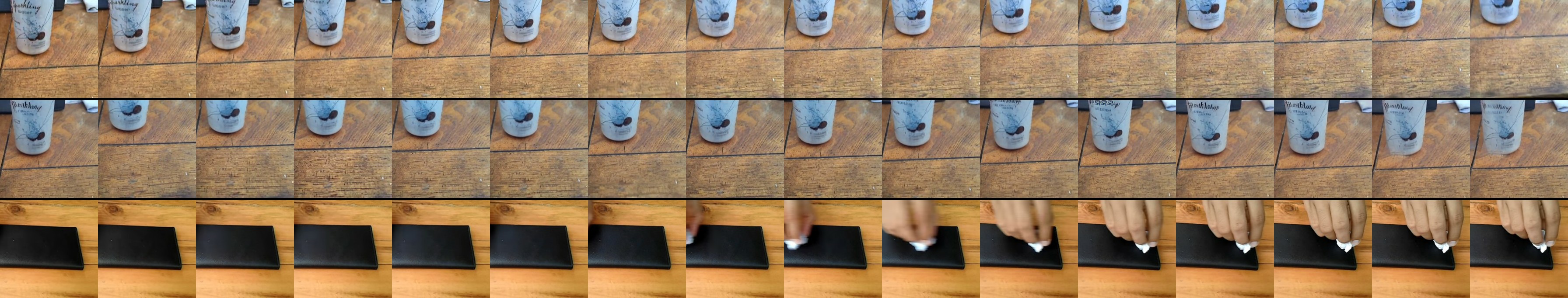}
    \caption{\textbf{Recombining the tokens of two videos} (\cref{sec:supp_probe}). Each block shows, from top to bottom, video A, which gives the primitive tokens of the first frame, the decoded video, and video B, which gives the derivative tokens of the later frames.}
    \label{fig:supp_mixed_gallery}
\end{figure*}

\begin{figure*}[t]
    \centering
    {\sffamily\let\small\footnotesize\input{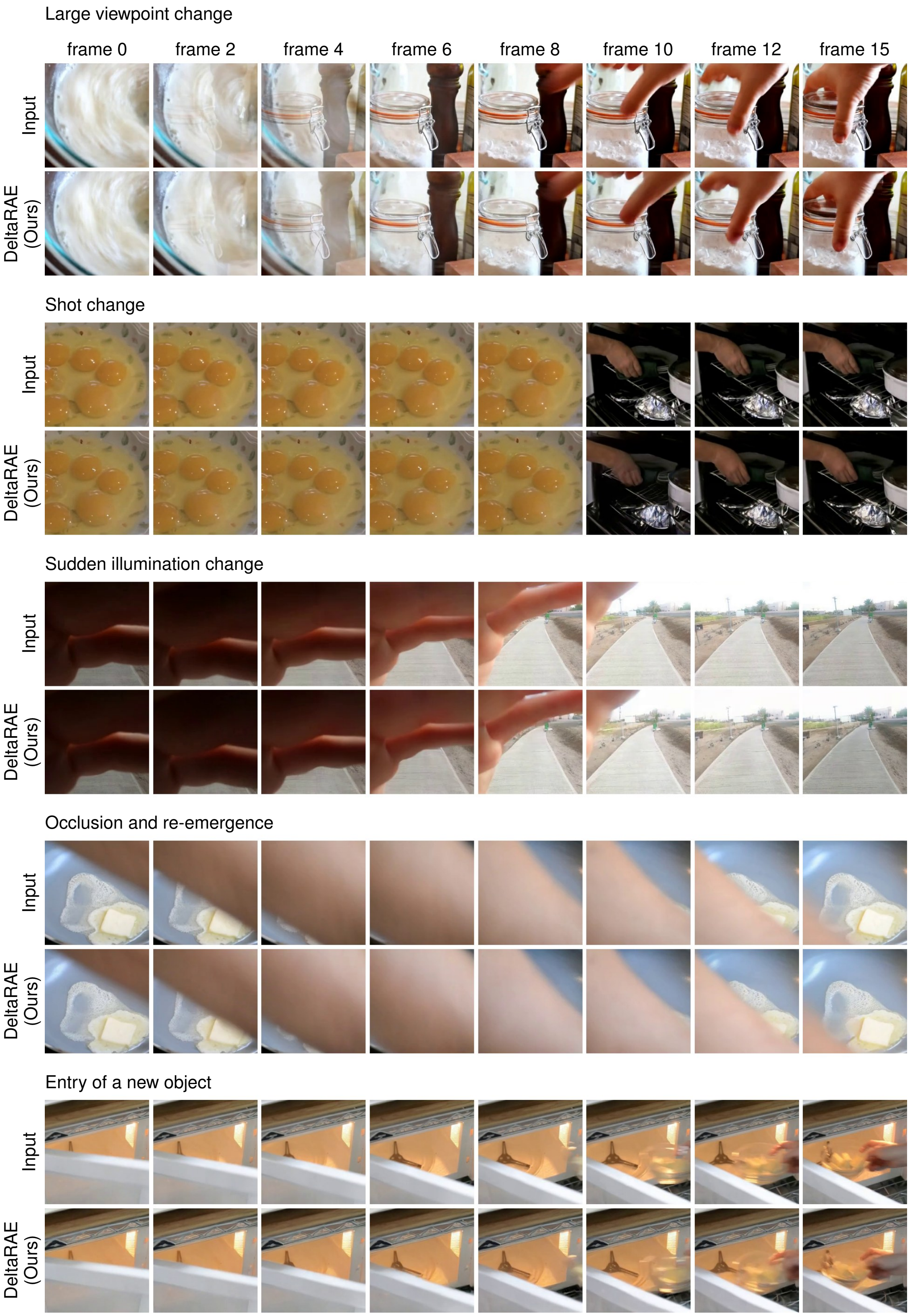}}
    \caption{\textbf{Difficult 16-frame clips reconstructed with 736 tokens.}}
    \label{fig:supp_edge_cases}
\end{figure*}

\cref{tab:long_recon,fig:supp_per_frame_curves} give the one-minute reconstruction numbers for all tokenizers and their evolution over time. \cref{fig:supp_long_recon} shows one-minute reconstructions, \cref{fig:supp_long_gen,fig:supp_long_gen_full} one-minute continuations, \cref{fig:supp_mixed_gallery} recombinations of the tokens of two videos, and \cref{fig:supp_edge_cases} difficult short clips.

\begin{table*}[t]
    \centering
    \small
    \caption{\textbf{Reconstruction of one-minute videos} on held-out Ego4D videos (\cref{sec:supp_protocol}). In the stitched rows, 16-frame tokenizers are applied clip by clip without finetuning. ``Drift'' is the slope of a linear fit to the PSNR of the 36 anchor periods; ``phase spread'' is the range of the mean PSNR over the four frame positions inside a group.}
    \label{tab:long_recon}
    \setlength{\tabcolsep}{4pt}
    \begin{tabularx}{\linewidth}{L c c | c c c c c}
        \toprule
        \textbf{Tokenizer} & \makecell[b]{\textbf{Tokens}\\\textbf{/ frame}} & \makecell[b]{\textbf{Tokens}\\\textbf{/ min}} & \textbf{PSNR} $\uparrow$ & \textbf{SSIM} $\uparrow$ & \textbf{LPIPS} $\downarrow$ & \makecell[b]{\textbf{Drift}\\\textbf{(dB/min)}} & \makecell[b]{\textbf{Phase spread}\\\textbf{(dB)} $\downarrow$} \\
        \midrule
        Dense per-frame grid & 256 & 368,640 & \multicolumn{5}{c}{\textit{not trained at this length}} \\
        16-frame \ourmodel\ (Ours), 1,216 tokens, stitched & 76.0 & 109,440 & 24.39 & 0.700 & 0.096 & $-0.19$ & 0.11 \\
        16-frame \ourmodel\ (Ours), 736 tokens, stitched & 46.0 & 66,240 & 23.42 & 0.668 & 0.113 & $-0.13$ & 0.29 \\
        \midrule
        w/o pretrained encoder & 13.6 & 19,584 & 23.29 & 0.673 & 0.175 & $-0.11$ & 3.92 \\
        \rowcolor{cvprblue!10}\ourmodel\ (Ours) & 13.6 & 19,584 & 19.84 & 0.568 & 0.213 & $-0.10$ & 0.73 \\
        \bottomrule
    \end{tabularx}
\end{table*}

\FloatBarrier
\section{Implementation and Hyperparameters}
\label{sec:supp_impl}

\paragraph{Attention at 1,440 frames.}
At 1,440 frames the interleaved encoder and decoder sequences have about 112,000 tokens, which dense attention cannot hold in memory. We implement the block-causal mask of \cref{eq:block_causal_mask} with sequence-parallel attention that skips the masked blocks between 240-frame chunks. Doubling the duration from 720 to 1,440 frames then doubles the step time, where dense attention would quadruple it. Videos are decoded at reduced resolution in the input pipeline to bound memory. \cref{tab:complexity_full_supp} reports encoder, decoder and total GFLOPs with measured latency for token budgets from 4,096 down to 496, together with the measured cost of one forward pass of the generator. From 4,096 to 736 tokens the cost of a forward pass falls by 87.5\%, a factor of 8.0 (\cref{tab:compute_cost}). The hyperparameters are listed in \cref{tab:arch_tok_supp,tab:arch_dit_supp}. Position is encoded in the same way in all tokenizers. A fixed 2D sine-cosine embedding is added to the patch features in the encoder and to the target tokens in the decoder; the register tokens receive no spatial embedding. A 1D rotary embedding~\cite{su2021roformer} over the frame index, or over the group index for one-minute models, is applied to the queries and keys of every attention layer of the encoder and the decoder, so attention depends on time only through the distance between groups. The generator uses learned embeddings for frame index, slot and token role instead.

\begin{table*}[t]
    \centering
    \small
    \caption{\textbf{Compute for token budgets from 4,096 down to 496} on 16 frames.}
    \label{tab:complexity_full_supp}
    \setlength{\tabcolsep}{4pt}
    \begin{tabular}{c c c | c c c | c c | c}
        \toprule
        \makecell[b]{\textbf{Derivative}\\\textbf{tokens / f}} & \makecell[b]{\textbf{Tokens}\\\textbf{/ 16f}} & \textbf{Compression} & \makecell[b]{\textbf{Enc.}\\\textbf{GFLOPs}} & \makecell[b]{\textbf{Dec.}\\\textbf{GFLOPs}} & \makecell[b]{\textbf{Total}\\\textbf{GFLOPs}} & \makecell[b]{\textbf{FLOPs}\\\textbf{speed-up}} & \makecell[b]{\textbf{Latency}\\\textbf{(speed-up)}} & \makecell[b]{\textbf{DiT GFLOPs}\\\textbf{/ pass}} \\
        \midrule
        256 & 4,096 & $1.00\times$ & 2,087.87 & 3,910.65 & 5,998.52 & $1.00\times$ & 13.41s ($1.00\times$) & 5,859 \\
        64 & 1,216 & $3.37\times$ & 1,408.93 & 1,963.33 & 3,372.26 & $1.78\times$ & 6.42s ($2.09\times$) & 1,282 \\
        \rowcolor{cvprblue!12}32 & 736 & $5.57\times$ & 1,316.00 & 1,699.16 & 3,015.16 & $1.99\times$ & 5.55s ($2.42\times$) & 730 \\
        16 & 496 & $8.26\times$ & 1,271.71 & 1,573.56 & 2,845.26 & $2.11\times$ & 5.31s ($2.53\times$) & --- \\
        \bottomrule
    \end{tabular}
\end{table*}

\section{Tokenizer Without the Pretrained Encoder}
\label{sec:supp_fr4}

To separate the effect of the frozen encoder from reconstruction quality, we train a tokenizer with the same token schedule and causal masks but without DINOv2, using pixel patches as encoder input. It reaches an rFVD similar to that of the DINOv2 tokenizer and about 6 dB higher PSNR (\cref{tab:semantic}). The generator does not benefit: K600 gFVD is about four to five times higher, and the gap remains when both encoders are paired with one DDPM generator recipe (\cref{tab:k600_gen_supp}). Matching rFVD therefore does not make a latent space suitable for generation, at 16 frames and, in the main paper, at one minute (\cref{tab:long_cap,fig:long_gen}). VideoRAE~\cite{xie2026videorae} reports a related observation across encoders, and ViTok~\cite{hansen2025vitok} across bottleneck sizes.

\begin{table*}[t]
    \small\centering
    \begin{minipage}[t]{0.46\textwidth}
    \centering
    \caption{\textbf{Frozen DINOv2 encoder versus a tokenizer without the pretrained encoder at matched rFVD} on K600; gFVD at $64^2$. The frozen-DINOv2 rows are the fully causal tokenizers with the pixel-residual stem (\cref{tab:causality}). These numbers are from the ablation setup, not from the final system (\cref{tab:main}).}
    \label{tab:semantic}
    \setlength{\tabcolsep}{6pt}
    \begin{tabular}{@{}l c c c@{}}
        \toprule
        \textbf{Tokenizer} & \textbf{rFVD} $\downarrow$ & \textbf{PSNR} $\uparrow$ & \textbf{gFVD} $\downarrow$ \\
        \midrule
        \multicolumn{4}{@{}l}{\emph{736 tokens}} \\
        Frozen DINOv2 & 5.77 & 24.10 & 5.7 \\
        w/o pretrained encoder & 3.44 & 30.01 & 22.7 \\
        \midrule
        \multicolumn{4}{@{}l}{\emph{1,216 tokens}} \\
        Frozen DINOv2 & 3.50 & 24.68 & 3.8 \\
        w/o pretrained encoder & 3.09 & 30.81 & 19.7 \\
        \bottomrule
    \end{tabular}
    \end{minipage}
\end{table*}

\begin{figure}[t]
    \centering
    \includegraphics[width=\linewidth]{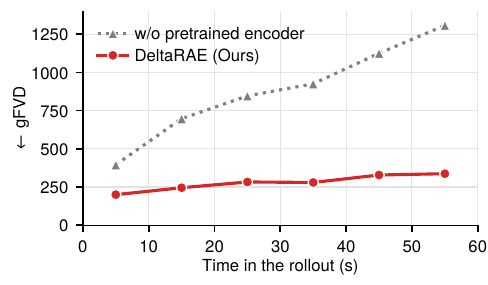}
    \caption{\textbf{gFVD of the one-minute rollouts over time,} with the same generator recipe for both tokenizers. gFVD is computed on 16-frame windows, 16 in each 10-second segment, for the rollouts of the 455 held-out videos. The native resolution is $224^2$, and $256^2$ without the pretrained encoder.}
    \label{fig:long_gen_fvd}
\end{figure}

\paragraph{gFVD of the one-minute rollouts.}
\cref{fig:long_gen_fvd} measures the rollouts of \cref{tab:long_cap} with gFVD on 16-frame windows. Over the minute \ourmodel reaches 231.5 and the tokenizer without the pretrained encoder 765.9. The gap between the two grows from a factor of 2.0 in the first 10 seconds to 3.9 in the last.

\section{Recombining Primitive and Derivative Tokens}
\label{sec:supp_probe}

\begin{figure*}[t]
    \centering
    {\sffamily\let\small\footnotesize\input{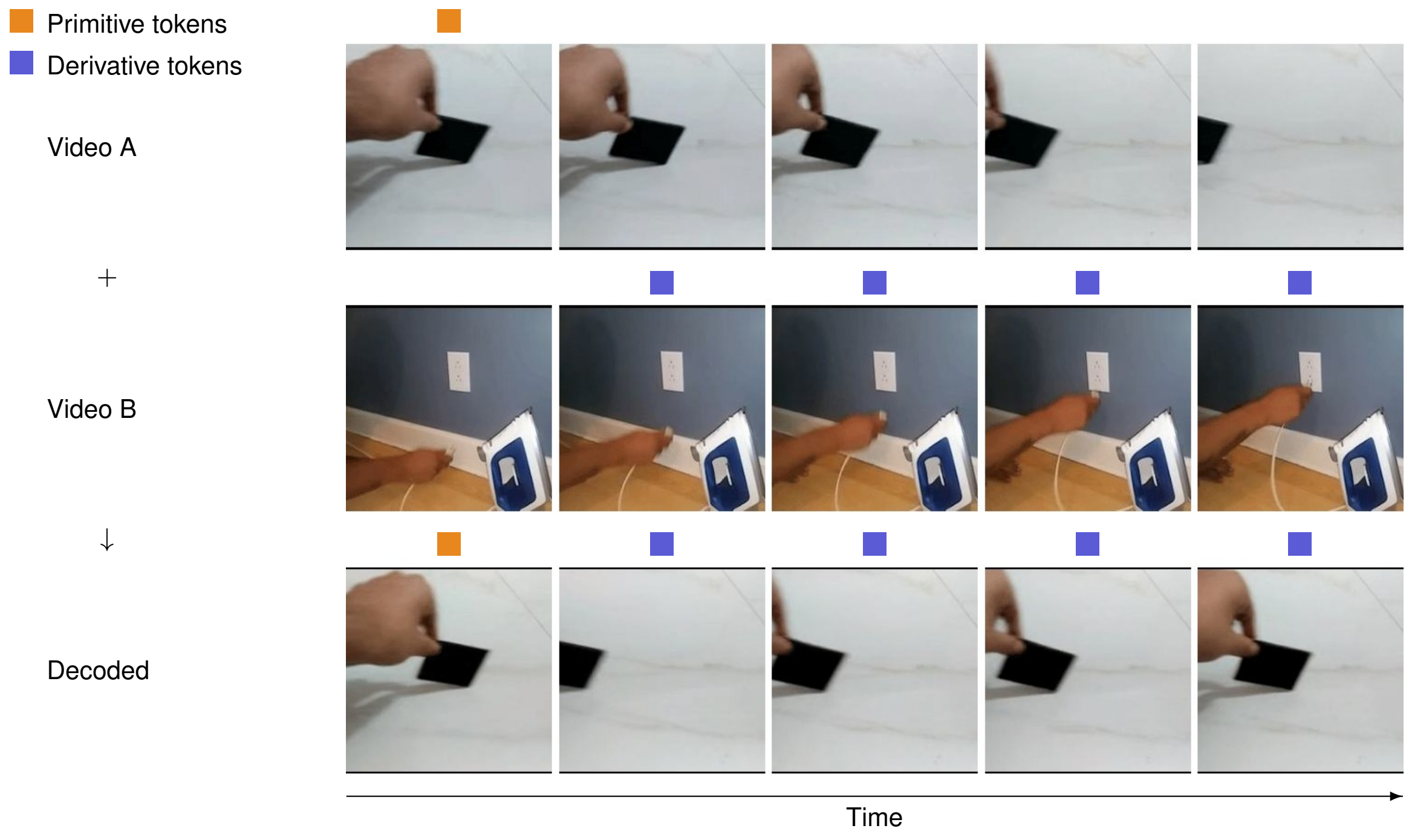}}
    \caption{\textbf{Recombining the tokens of two videos.} The primitive tokens of the first frame of video A and the derivative tokens of the later frames of video B are decoded together, without finetuning. Squares mark the frames whose tokens are used.}
    \label{fig:motion_disentanglement}
\end{figure*}

The split into primitive and derivative tokens raises the idea that a video tokenizer could keep appearance and change in separate tokens. This section is a first probe of that idea. We decode the primitive tokens of one video together with the derivative tokens of another, without finetuning (\cref{fig:motion_disentanglement}). The output keeps the appearance of the video that supplied the primitive tokens: over 100 random Kinetics-600 pairs, CLIP~\cite{radford2021learning} similarity is 0.932 to that video and 0.540 to the other, compared with 0.528 for unrelated pairs. This is consistent with derivative tokens carrying little appearance. We draw no conclusion on whether they carry the motion of the other video, for which there is no agreed metric.

\clearpage
{\small
}

\end{document}